\documentclass[preprint, 1p]{elsarticle}
\usepackage{natbib}
\usepackage{amsmath,amssymb,amsfonts}
\usepackage{algorithmic}
\usepackage{graphicx}
\usepackage{hyperref}
\usepackage{amsthm}
\usepackage{tabularx}
\usepackage{array}
\usepackage{multirow}
\usepackage{booktabs}
\usepackage[ruled,linesnumbered,vlined]{algorithm2e}
\usepackage{verbatim}
\usepackage[figuresright]{rotating}
\usepackage{subfigure}
\usepackage[labelfont=bf,textfont={bf}]{caption}
\usepackage{xcolor}
\usepackage{bbding}

\usepackage{textcomp}
\usepackage{makecell}

\usepackage{enumitem}

\newcommand{\model}{AR-DBSCAN}
\newcommand{\mypara}[1]{{\vspace{1.5mm}\noindent\textbf{#1}}}

\makeatletter
\newcommand{\algorithmfootnote}[2][\footnotesize]{%
  \let\old@algocf@finish\@algocf@finish
  \def\@algocf@finish{\old@algocf@finish
    \leavevmode\rlap{\begin{minipage}{\linewidth}
    #1#2
    \end{minipage}}%
  }%
}
\makeatother

\journal{Artificial Intelligence }

\begin{document}

\begin{frontmatter}

\title{Adaptive and Robust DBSCAN with Multi-agent Reinforcement Learning}


\author[label]{Hao Peng\corref{cor1}}
\ead{penghao@buaa.edu.cn}
\cortext[cor1]{Corresponding author.}
\author[label]{Xiang Huang}
\author[label]{Shuo Sun}
\author[label]{Ruitong Zhang}
\author[label2]{Philip S. Yu}
\affiliation[label]{organization={Beihang University},
            country={China}}
\affiliation[label2]{organization={University of Illinois Chicago},
            country={US}}

\begin{abstract}
DBSCAN, a well-known density-based clustering algorithm, has gained widespread popularity and usage due to its effectiveness in identifying clusters of arbitrary shapes and handling noisy data. 
However, it encounters challenges in producing satisfactory cluster results when confronted with datasets of varying density scales, a common scenario in real-world applications.
In this paper, we propose a novel \textbf{A}daptive and \textbf{R}obust \textbf{DBSCAN} with Multi-agent Reinforcement Learning cluster framework, namely \textbf{\model{}}. 
First, we model the initial dataset as a two-level encoding tree and categorize the data vertices into distinct density partitions according to the \emph{information uncertainty} determined in the encoding tree.
Each partition is then assigned to an agent to find the best clustering parameters without manual assistance. 
The allocation is density-adaptive, enabling \model{} to effectively handle diverse density distributions within the dataset by utilizing distinct agents for different partitions.
Second, a multi-agent deep reinforcement learning guided automatic parameter searching process is designed.
The process of adjusting the parameter search direction by perceiving the clustering environment is modeled as a Markov decision process.
Using a weakly-supervised reward training policy network, each agent adaptively learns the optimal clustering parameters by interacting with the clusters.
Third, a recursive search mechanism adaptable to the data's scale is presented, enabling efficient and controlled exploration of large parameter spaces.
Extensive experiments are conducted on nine artificial datasets and a real-world dataset. 
The results of offline and online tasks show that \model{} not only improves clustering accuracy by up to 144.1\% and 175.3\% in the NMI and ARI metrics, respectively, but also is capable of robustly finding dominant parameters.
\end{abstract}


\end{frontmatter}

\section{Introduction}\label{sec:introduction}
Clustering serves as a fundamental technology for advancing machine intelligence~\cite{10.1145/3626772.3657931, 10.1145/3626772.3657972, 10.1145/3543507.3583362}.
It reveals the intrinsic structure of data and aids machine intelligence systems in organizing and comprehending information.
Clustering algorithms aim to divide data into multiple clusters based on specific rules.
By grouping similar samples (defined by similar distance measurement) into the same cluster and allocating dissimilar samples into different ones, these algorithms reveal the differences between samples and the underlying patterns connecting them.
As a typical density-based clustering method, Density-Based Spatial Clustering of Applications with Noise (DBSCAN)~\cite{ester1996density} defines clusters as the maximal set of density-connected points. 
It partitions regions of sufficient density into different clusters and identifies clusters with arbitrary shapes in spatial datasets with noise.
Due to its simplicity and practicality, DBSCAN has gained widespread recognition and is extensively used in various scientific and engineering fields, such as image segmentation~\cite{hou2016dsets}, commercial research~\cite{schubert2017dbscan}, biological analysis~\cite{wang2020theoretically}, neuroscience~\cite{mai2012similarity}, target recognition~\cite{zhu2022webface260m}, and more~\cite{mai2020incremental}.

\begin{figure*}
\centering
\includegraphics[width=0.8\textwidth]{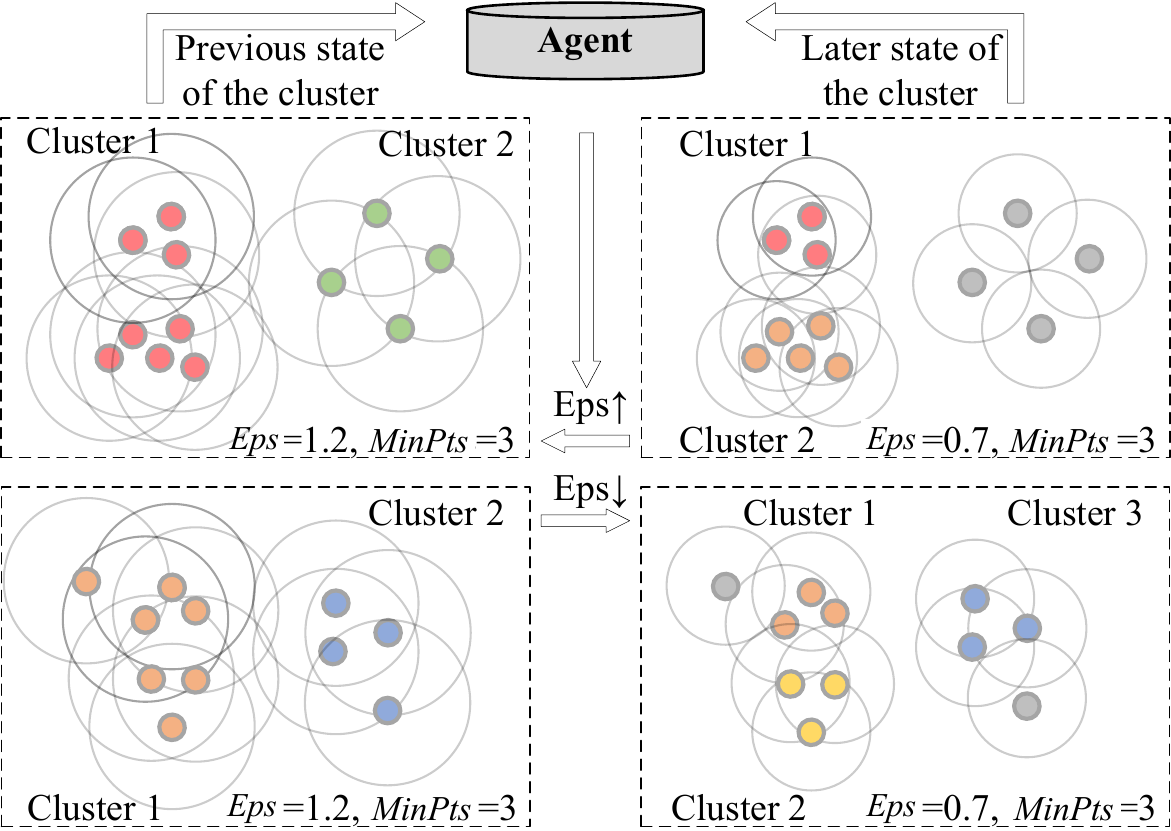}
\caption{The challenge of parameter searching through a single agent. Finding the optimal parameter combination of $Eps$ and $MinPts$ in complex situations characterized by disparate density in the data is difficult. \textmd{The gray vertices depict unclustered noise points.}}
\label{fig1}
\end{figure*}

However, the two predefined parameters of DBSCAN, i.e., the distance of the cluster formation $Eps$ and the minimum data objects required inside the cluster $MinPts$, which typically need to be determined manually, pose four challenges to the clustering process.
\textbf{First, density variety challenge.} 
When the data density distribution is not uniform, and the intra-cluster distance varies significantly, DBSCAN with only one parameter combination can hardly perform well. 
In the scenario depicted in Figure~\ref{fig1}, certain intra-cluster distances are small while others are large, eg., \emph{Cluster 1} and \emph{Cluster 2} in the upper left subfigure of Figure~\ref{fig1}. 
When determining the value of $Eps$ based on a smaller cluster scale, points with slightly larger distances may be classified as noise points incorrectly (the upper half of Figure~\ref{fig1}).
Conversely, if $Eps$ is set too high, different clusters may be erroneously merged into a single cluster (the bottom half of Figure~\ref{fig1}).
Some researchers have attempted to address this issue~\cite{kim2019aa, wang2015adaptive, wang2022amd} by defining parameters for different groups based on varying densities, but they require introducing specific prior information, such as the total number of clusters or the value of $MinPts$.
Therefore, the first challenge lies in devising a method to achieve accurate clustering when clusters have varying densities.
\textbf{Second, parameters free challenge.} 
The clustering results are greatly subject to the two parameters $Eps$ and $MinPts$, but they must be determined beforehand based on practical experience.
Existing methods~\cite{liu2007vdbscan, mitra2011kddclus, wang2022amd} based on the $k$-distance estimate the possible values of the $Eps$ through significant changes in the curve of the k-dist plot, but it still requires manual determination of the $MinPts$ parameter beforehand.
Although some improved DBSCAN methods avoid the simultaneous adjustment of $Eps$ and $MinPts$ to varying degrees, they also need to pre-determine the cutoff distance parameter~\cite{diao2018improved}, grid parameter~\cite{chang2008robust}, Gaussian distribution parameter~\cite{smiti2012dbscan}, fixed $MinPts$ parameter~\cite{akbari2016automated, hou2016dsets}, or $k$-NN parameter and relative density threshold parameter~\cite{QIAN2024127329}.
Therefore, the second challenge is to automatically find optimal cluster parameters for DBSCAN without the need for manual tuning based on expert knowledge.
\textbf{Third, adaptive policy challenge.} 
Traditional parameter search methods for DBSCAN, which rely on fixed patterns~\cite{mitra2011kddclus, liu2007vdbscan, BAT-DBSCAN}, often face bottlenecks when dealing with unconventional data problems. 
This is due to the varying data distributions and cluster characteristics in clustering tasks.
Even some hyperparameter optimization methods~\cite{10.5555/2986459.2986743, karami2014Choosing, lessmann2005optimizing} that utilize an external clustering evaluation index based on label information as the objective function are ineffective when data label information is absent.
On the other hand, approaches~\cite{8777081, li2007discovery} that only use the internal clustering evaluation index as the objective function are limited by accuracy issues, even though they do not require label information.
Therefore, considering the absence of label information, the third challenge is effectively and adaptively adjusting the parameter search policy for the data in each process.
\textbf{Fourth, search efficiency challenge.}
The parameter space expands as the data scale increases.
However, an excessively large search space hinders search efficiency~\cite{kanervisto2020action} and introduces noise that affects clustering accuracy~\cite{dulac2015deep}.
Thus, the fourth challenge lies in enhancing search efficiency while upholding high clustering accuracy.

In recent years, the theory of structural entropy~\cite{li2016structural} has demonstrated significant advantages in graph-structured data analysis~\cite{li2018decoding, liu2019rem, wu2022structural, zou2023se, huang2024sec}. 
An abstracting partitioning tree, also named \emph{encoding tree}, is generated from a graph by minimizing structural entropy, where each node in the encoding tree has its corresponding node structural entropy.
The structural entropy of the whole graph is calculated by summing up all non-root nodes' structural entropy in the encoding tree\footnote{Vertices are defined in the graph, and nodes are in the tree.}.
Meanwhile, by aggregating the structural entropy of the leaf nodes to the intermediate nodes within the encoding tree, it becomes feasible to quantify the density distribution, as partitions with similar densities exhibit similar structural entropy.
On the other hand, deep reinforcement learning~\cite{fujimoto2018addressing, lillicrap2015continuous} also has achieved remarkable success in tasks lacking training data due to its ability to learn by receiving environmental feedback.
By engaging in continuous interactions with the environment, it acquires the optimal policy in different scenarios.

To tackle the challenges mentioned above, we propose a novel \textbf{A}daptive and \textbf{R}obust \textbf{DBSCAN} with Multi-agent Reinforcement Learning cluster framework, namely \textbf{\model{}}. 
The framework contains two methods: \emph{agent allocation} and \emph{automatic parameter search}, as well as an accelerated \emph{recursive search mechanism}.
First, we design a new structural entropy based agent allocation method, which partitions the initial dataset based on the \emph{information uncertainty distribution} of intermediate nodes on the encoding tree and allocates an agent for each partition. 
A new considering data scale normalization based structured graph construction algorithm is given to convert initial discrete data into a graph, capturing neighborhood information for each data vertice.
Unlike the previous one-dimensional structural entropy maximization based $k$-NN structuralization approach~\cite{li2016three}, the new algorithm considers mitigating the impact of data scale and the variation in $k$ values of the $k$-NN graph.
Besides, a new information uncertainty metric is devised to measure the density distribution of data partitions associated with the intermediary nodes.
Second, we present a new multi-agent deep reinforcement learning based automatic parameter search method to find the optimal parameter combination for each partition. 
The parameter search process for each agent is transferred into a Markov Decision Process where the cluster changes observed after each clustering step serve as the state, and the parameter adjustment acts as the action. 
It is important to note that the agents operate in a non-competitive manner.
Through weak supervision, we construct rewards based on a limited number of external clustering indices. 
Additionally, we employ a new attention mechanism by fusing global and local states of multiple clusters, prompting the agents to attend to the global search environment and allocate varying levels of attention to distinct clusters.
Third, a new personalized recursive search mechanism for each agent is devised to gradually reduce the search range and improve the precision of the search.
The data scale determines the maximum searching parameter boundaries, and the variations in search range and search precision are controllable. 

Comprehensive experiments are carried out to evaluate the clustering performance of \model{} on nine commonly used clustering datasets and one online task dataset.
The results indicate that compared with the state-of-the-art parameter search approaches, \model{} exhibits substantial improvements in accuracy, with enhancements of up to 144.1\% and 175.3\% in NMI and ARI metrics, respectively, demonstrating its effectiveness without requiring manual assistance.
Moreover, \model{} significantly enhances the robustness of clustering, with a reduction in the variance of up to 0.442 and 0.463 in the NMI and ARI metrics, respectively.
All codes and datasets of this work are publicly available at GitHub\footnote{\url{https://github.com/RingBDStack/AR-DBSCAN}}.

Our preliminary work appeared at the proceedings of the ACM International Conference on Information and Knowledge Management 2022~\cite{zhang2022automating} and won the Honorable Mention for Best Paper\footnote{http://www.cikmconference.org/2022/best-paper.html}.
The journal version in this paper has extended the original parameter search framework into an adaptive and robust parameter search framework with multi-agent reinforcement learning. 
This full version involves several improvements in upgrading the methodology and the framework structure of the proposed architecture. 
In terms of model upgrades, different from learning parameters for the whole dataset by one agent, we introduce a multi-agent parameter search framework to learn adaptive parameters for different density partitions to improve the accuracy of clustering in complex datasets with various cluster densities. 
We integrate structural entropy theory into our framework to provide a view of the density distribution of the data, which guides the agent allocation.
A multi-agent recursive mechanism is presented to personalize the adjustment of parameter search space and precision for various density partitions.
More state-of-the-art performances are presented and analyzed in the experiments.
Comprehensive experiments on ten datasets show the effectiveness and robustness of~\model{}.

In summary, the contributions of this paper are summarized as follows:

\noindent$\bullet$ A new parameter search framework guided by multi-agent deep reinforcement learning, named \model{}, is presented. 
\model{} achieves adaptive and robust clustering on datasets with varying densities, eliminating the requirement for pre-determining DBSCAN parameters.

\noindent$\bullet$ A novel agent assignment method, based on structural entropy, is designed to assign agents by considering the information uncertainty distribution of intermediate nodes in the encoding tree. The newly proposed information uncertainty metric provides a new perspective for quantifying the density distribution of data partitions.

\noindent$\bullet$ A new non-competitive multi-agent parameter search framework is devised to adaptively explore the optimal combination of parameters for partitions. This framework utilizes weak supervision and attention mechanism capturing both global information and local information within clusters.

\noindent$\bullet$ A new personalized recursive parameter search mechanism is introduced, enabling efficient and controlled search for all agents.

\noindent$\bullet$ Comprehensive experiments and analyses are conducted on ten datasets, demonstrating that \model{} outperforms the existing state-of-the-art approaches in terms of accuracy, efficiency, and robustness.

This paper is organized as follows:
Sec.~\ref{sec:prolim} outlines the problem formulation and notation descriptions, 
Sec.~\ref{sec:model} describes the detailed designs of \model{}, 
Sec.~\ref{sec:setup} showcases the datasets and experiment details,
Sec.~\ref{sec:evaluation} gives the results and analyses,
and Sec.~\ref{sec:relate} introduces the related work before our conclusion in Sec.~\ref{sec:conclution}.

\section{Problem formulation and notation}\label{sec:prolim}
In this section, we give the formula definitions of DBSCAN clustering, parameter search of DBSCAN, and corresponding concepts in the structural entropy theory~\cite{li2016structural}.
The forms and description of all necessary notations throughout our work are stated in Table~\ref{tab:notations}.

\noindent
\textbf{Defination 2.1. (DBSCAN clustering).}
Given data vertices $V=\{v_1, ... , v_j, v_{j+1}, ...\}$, DBSCAN aims to find clusters $C=\{c_1, ..., c_n, c_{n+1}, ...\}$ based on the parameter combination $P=\{Eps, MinPts\}$. 
Here, $Eps$ means the maximum distance that two objects can be formed in a cluster, and $MinPts$ measures the minimum number of adjacent objects where a core object can reach within $Eps$ distance.
Concretely, DBSCAN first selects a random point having at least $MinPts$ points within $Eps$ distance. 
Then, each point within the neighborhood of the core point is evaluated to see if it has $MinPts$ points within $Eps$ distance (including the point itself). 
If the point satisfies the $MinPts$ criterion, it becomes another core point, and the cluster expands. 
If a point does not satisfy the $MinPts$ criterion, it becomes a boundary point.
And a point is classified as noise if it does not meet the criteria to be considered a core point or a boundary point~\cite{ester1996density}.

\noindent
\textbf{Defination 2.2. (Parameter search of DBSCAN).}
Given the partition set $\{\mathcal{P}_1, \mathcal{P}_2, \dots, \mathcal{P}_t, \dots\}$, for the $t$-th partition $\mathcal{P}_t = \{v_1^t, \dots, v_i^t, v_j^t, \dots\}$, the parameter search of DBSCAN clustering is the process of finding the optimal parameter combination $P_t=\{Eps_t, MinPts_t\}$ in corresponding parameter spaces.

\noindent
\textbf{Defination 2.3. (Parameter search of DBSCAN in the online task).}
Given the data blocks $\{\mathcal{V}_1, \dots, \mathcal{V}_i, \mathcal{V}_{i+1}, \dots\}$ as the online data stream, the parameter search in online clustering is the process of finding the optimal parameter combinations for each data block $\mathcal{V}_i$.

\noindent
\textbf{Defination 2.4. (Structured Graph).}
Let $G=\{V, E, W\}$ denote as a weighted undirected graph, where $V$ is the vertices set, $E$ is the edge set and $W\in \mathbb{R}^{+}$ is the edge weight set. 
The degree of vertice $v$, denoted as $d_v$, is defined as the sum of associated edge weights.

\noindent
\textbf{Defination 2.5. (Encoding Tree).}
Given a graph $G=\{V, E, W\}$, the encoding tree $T$ of $G$ is defined with the following properties: 
\textbf{(1)} For the root node denoted $\lambda$, we define the vertices set it associated $T_\lambda=V$. 
\textbf{(2)} For every node $\alpha \in T$, the immediate successors of $\alpha$ are $\alpha^{\left \langle j \right \rangle}$ ordered from left to right as $j$ increases. 
It is required that every internal node has at least two immediate successors.
\textbf{(3)} For each non-root node $\alpha$, its parent node in $T$ is denoted as $\alpha^-$.
\textbf{(4)} For node $\alpha \in T$, the vertex subsets $T_\alpha=\bigcup_{i=1}^{N_\alpha}T_{\alpha^{\left \langle j \right \rangle}}$, $N_\alpha$ is number of successors of $\alpha$. 
We call an encoding tree the $K$-level encoding tree if its height is $K$.

\begin{table*}[t]
\aboverulesep=0ex
\belowrulesep=0ex
    \caption{Forms and interpretations of notations.}
    \centering
    \resizebox{1.2\columnwidth}{!}{
    \begin{tabular}{l|l}
        \toprule
        \textbf{Notation} & \textbf{Description}\\
        \hline
        $C$; $c_{n}$; $\mathcal{P}$; $\mathcal{P}_i$ &Cluster set; The $n$-th cluster; The data partition; The $i$-th data partiton\\
        $Eps$; $MinPts$ & The distance of the cluster; The minimum data required inside the cluster\\
        $P$ & The parameter combination for DBSCAN\\
        \hline
        $G$; $V$; $E$; $v$; $d_v$ & The graph; The vertices set; The edges set; The data vertice; The degree of $v$\\
        $e_{i,j}$; $W$; $w_{i, j}$ & The edge between $v_i$ and $v_j$; The edge weight set; The weight of $e_{i,j}$\\
        $vol(G)$ & The volume of graph $G$, i.e., degree sum in $G$\\
        $T$; $\lambda$; $\alpha$& The encoding tree; The root node of $T$; The node of $T$\\
        $\alpha^{-}$;$\alpha^{\left \langle j \right \rangle}$ & The parent node of $\alpha$; The $j$-th child of $\alpha$ \\
        $N_{\alpha}$; $T_{\alpha}$ & The number of immediate successor nodes of $\alpha$; The vertices set corresponding to node $\alpha$\\
        $H^{1}(G)$; $H^{1}_{norm.}(G)$ & The one-dimensional structural entropy of $G$; The normalized one-dim. structural entropy\\
        $H^{K}(G)$; $H^{T}(G)$ & The $K$-dimensional structural entropy of $G$; The structural entropy of $G$ under $T$\\
        $H^{T}(G;\alpha)$; $\mathcal{H}^{T}(G;\alpha)$ & The structural entropy of node $\alpha$; The normalized structural entropy of node $\alpha$\\
        \hline
        $D_{ij}$; $\Vert \cdot{} \Vert$ & The distance of vertice $v_i$ and $v_j$; The L2 norm function.\\
        $M(T;\alpha,\gamma)$;$C(T;\alpha,\gamma)$& The merging operator; The combining operator\\
        \hline
        $\mathcal{S}$; $s$; $\mathcal{A}$; $a$  & The state set; The state for one step; The action space; the action for one step\\
        $\mathcal{R}$; $r$; $\mathbf{P}$ & The reward function; The reward for one step; The policy optimization algorithm\\
        $\mathcal{V}$; $\mathcal{V}_{i}$; $|\cdot{}|$; $\cdot{}'$ & The data block; The $i$-th data block; The size of $\cdot{}$; The partial set of $\cdot{}$\\
        $e$; $I$ & The episode of parameter search process; The maximum step limit in an episode\\
        $\cdot{}^{(e)(i)}$; $s_{global}$; $s_{local, n}$ & The value of $\cdot{}$ in the $i$-th step of episode $e$; The global state; The local state of cluster $c_n$\\
        $\mathcal{D}_{b}$; $R_{cn}$ & The quaternary distances of $P$; The ratio of $|\mathcal{C}|$ and $\mathcal{V}$ \\
        $\mathcal{X}$; $\mathcal{Y}$; $\cdot_{center, n}$ & The feature set; The true label set; The value of $\cdot$ for the center object of $c_{n}$\\
        $F_{G}$; $F_{L}$ & The FCN for $s_{global}$; The FCN for $s_{local, n}$\\
        $F_{S}$; $\alpha_{att, n}$; $\sigma(\cdot)$  & The scoring FCN; The attention weight for cluster $c_{n}$; The \texttt{ReLU} activation function\\
        $\mathbin\Vert$; $\beta$, $\delta$ & The splicing operation; The reward impact factors\\
        $\mathcal{T}$; $M$; $\mathcal{L}_c$; $\mathcal{L}_a$ & The core element of one step; The sample number for $\mathcal{T}$; The loss for $Critic$; The loss for $Actor$\\
        \hline
        $d$; $P_o$, $Eps_o$, $MinPts_o$ & The dimension of feature; The optimal parameters\\
        $\cdot{}^{(l)}$ & The value of $\cdot{}$ in layer $l$\\
        $B_{p, 1}$; $B_{p, 2}$ & The minimum boundaries for parameter $p$; The maximum boundaries for parameter $p$\\
        $\pi_{p}$; $\theta_{p}$; $\lfloor\ \rfloor$ & The number of valid parameter; The parameter search step; The rounding down operation\\
        \bottomrule
    \end{tabular}
    }
    \label{tab:notations}
\end{table*}

\noindent
\textbf{Defination 2.6. (One-dimensional Structural Entropy).}
In a single-level encoding tree $T$, the one-dimensional structural entropy is defined as follows:
\begin{equation}
    H^1(G)=-\sum_{v \in V} {\frac{d_v}{vol(G)}} \cdot \log_2{\frac{d_v}{vol(G)}}, \label{eq:1-dim se}
\end{equation}
where $d_v$ is the degree of vertex $v$, and $vol(G)$ is the sum of the degrees of all vertices in $G$. 
As noted in previous research~\cite{li2016structural}, the one-dimensional structure represents the maximum level of information embedding in graph $G$.

\noindent
\textbf{Defination 2.7. (High-dimensional Structural Entropy).}
For the encoding tree $T$, the high-dimensional structural entropy of $G$ is defined as follows:
\begin{equation}
    H^K(G)=\mathop{\min}_{\forall T:height(T) \leq K}{H^T(G)}, \label{eq:k-dim se}
\end{equation}
\begin{equation}
    H^T(G)=\sum_{\alpha \in T, \alpha \neq \lambda} H^T(G;\alpha). \label{eq:tree se}
\end{equation}
Here $H^{T}(G;\alpha)$ is the structural entropy of node $\alpha$, and $H^{T}(G)$ is the structural entropy of encoding tree $T$. 
$H^K(G)$ is the $K$-dimensional structural entropy, with the optimal encoding tree of $K$-level.

\noindent
\textbf{Defination 2.8. (Structual Entropy of tree node $\alpha$).}
Given an encoding tree $T$, and non-root node $\alpha$ in $T$, the structural entropy of node $\alpha$ is defined as follows:
\begin{equation}
    H^T(G;\alpha)=-\frac{g_\alpha}{vol(G)} \log_2{\frac{V_\alpha}{V_{\alpha^-}}}.\label{eq:tree node se}
\end{equation}
Here $g_\alpha$ represents the total weight of the cut edges linking the vertices in $T_\alpha$ with those in the complement set $T_{\lambda}/T_{\alpha}$, where the latter comprises all vertices that are outside of $T_\alpha$.
$V_\alpha$ is the sum of degrees of all vertices in $T_\alpha$.

\section{\model{} framework}\label{sec:model}

\begin{figure*}[t]
\centering
\includegraphics[width=0.8\textwidth]{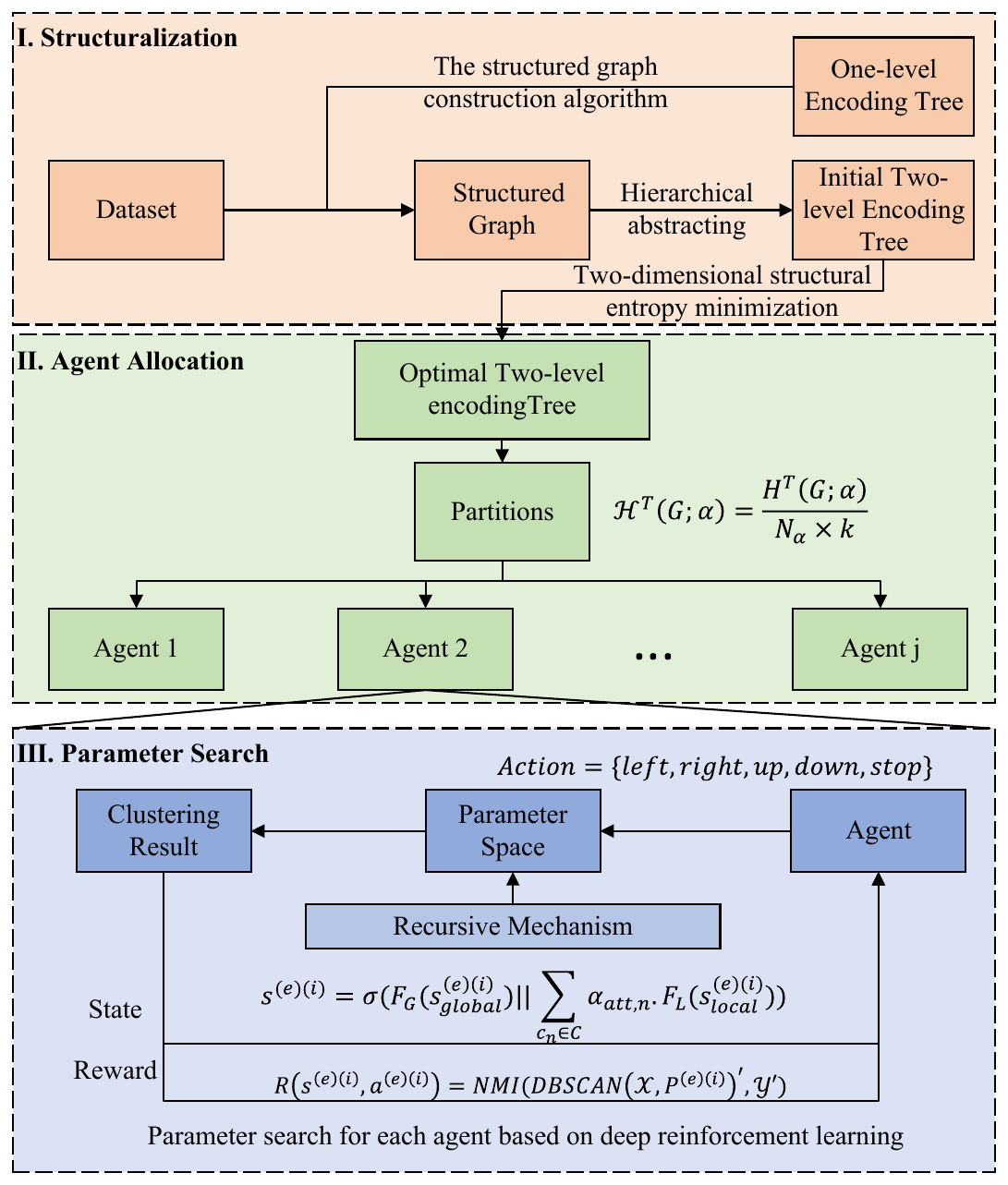}
\caption{Overall framework of \model{}.}
\label{fig:framework}
\end{figure*}

As illustrated in Fig.~\ref{fig:framework}, the details of \model{} consist of \emph{Structuralization}, \emph{Agent allocation}, and \emph{Parameter search}.
(1) \emph{Structuralization}. 
To begin with, the initial dataset is converted into a structural $k$-NN graph $G$, which captures the neighborhood structure information (Sec.~\ref{subsubsec:Structuredd graph construction}).
(2) \emph{Agent allocation}. 
Following the structuralization step, the optimal two-level encoding tree is generated from the graph $G$ based on two-dimensional structural entropy minimization (Sec.~\ref{subsubsec:Two-dimensional encoding tree optimization}).
The agent assignment method is then employed to partition the initial dataset based on information uncertainty distribution on the encoding tree. 
Each partition is assigned a corresponding agent (Sec.~\ref{subsubsec:agent assignment strategy}).
(3) \emph{Parameter search}. 
For each agent and corresponding data partitions, the separated parameter search process based on deep reinforcement learning is applied to find the optimal parameter combination respectively (Sec.~\ref{subsubsec:parameter search with DRL}).
The recursive mechanism is introduced to improve search efficiency in Sec.~\ref{subsubsec: Parameter Space and Recursion Mechanism}.
The time complexity of \model{} is given in Sec.~\ref{subsec:time complexity}.

\subsection{Structural entropy based agent allocation}\label{subsec:Structural entropy based agent allocation}
\begin{figure*}[t]
\centering
\includegraphics[width=0.8\textwidth]{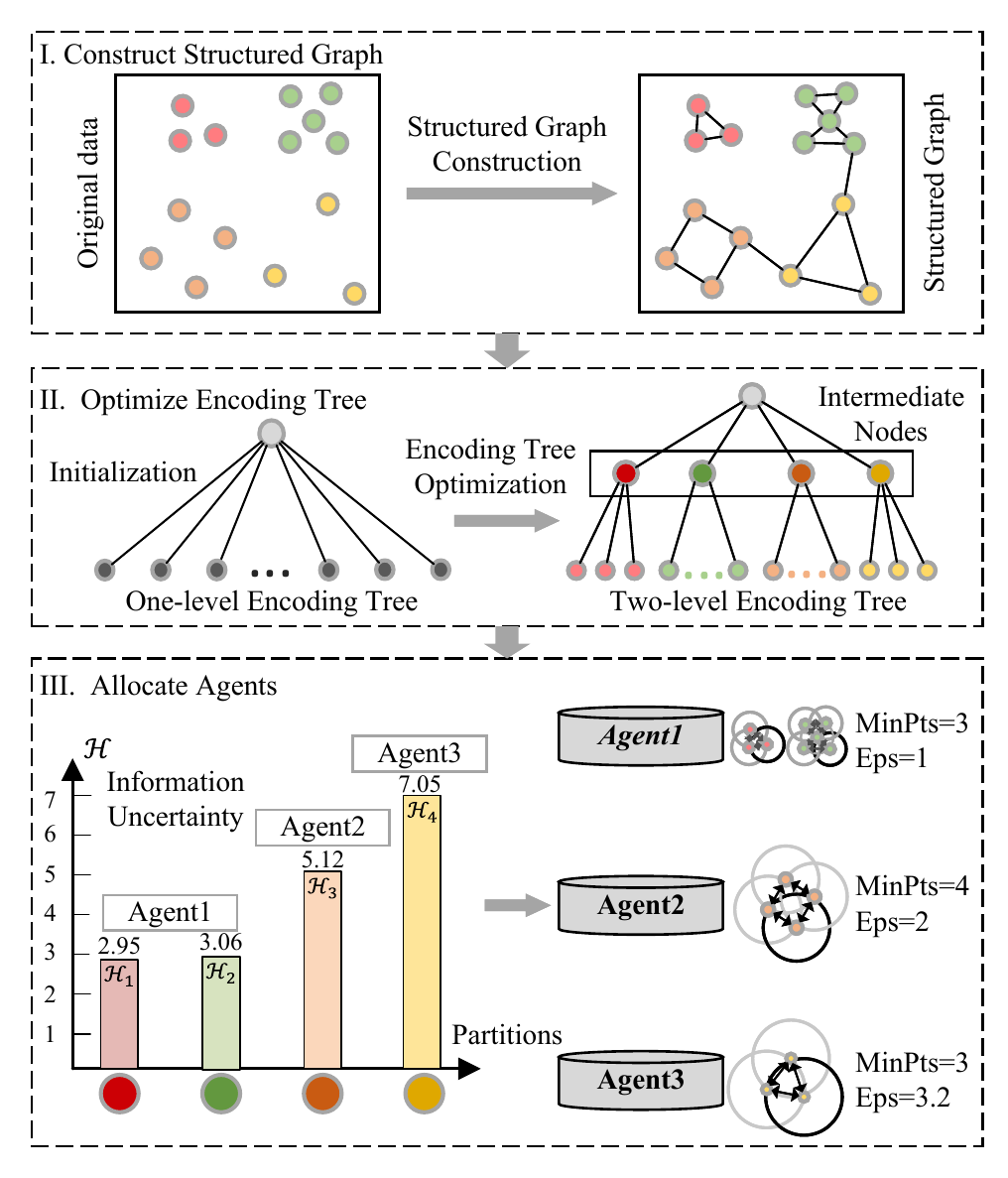}
\caption{Structural entropy based agent allocation.}
\label{fig:allocate-agent}
\end{figure*}

When dealing with clusters of varying densities, existing parameter search policies, such as those proposed in references~\cite{bergstra2012random, liu2007vdbscan, zhang2022automating}, which rely on a single parameter combination, may fail to accurately recognize them (as discussed in Sec.~\ref{subsubsec:offline acc} and Sec.~\ref{subsec:online evaluation}).
To tackle the collision of various density clusters in parameter search, a new structural entropy based agent allocation method is presented in Figure~\ref{fig:allocate-agent}. 

\subsubsection{Structured graph construction} \label{subsubsec:Structuredd graph construction}
We transfer the initial dataset to a structured graph $G=\{V, E, W\}$, where $V$ is the set of data points and $E$ is determined through a $k$-nearest neighbors ($k$-NN) approach employing Euclidean distance. 
The Euclidean distance is defined as follows:
\begin{equation}
    D_{ij}= \Vert r_i- r_j \Vert_{2} \label{eq:5 euclidean distance},
\end{equation}
where $r_i$ and $r_j$ are the coordinates or feature vectors of two vertices $v_i$ and $v_j$, and $\Vert \cdot{} \Vert_{2}$ is the L2 norm. 
The edge weight $W$, also referring to the similarity, is based on the Euclidean distance with normalization on the number of edges and the sum of distances. 
Specifically, the edge weight between vertices $v_i$ and $v_j$ is calculated as follows:
\begin{equation}
    \label{eq:edge weight}
    w_{i,j} = \exp(-D_{ij} \cdot \frac{|E|}{\sum_{e_{m, n}\in E} D_{m, n}}).
\end{equation}
Here $e_{m, n}$ is the edge in the edge set $E$, $D_{m, n}$ is the Euclidean distance between vertices $v_{m}$ and $v_{n}$, and $|E|$ is the size of the edge set.
The edge weight (similarity) is inversely proportional to the Euclidean distance.
Intuitively, the similarity increases as the distance between vertices decreases, leading to a larger edge weight.

Selecting the core parameter $k$ in $k$-NN is a critical challenge.
A larger $k$ may create excessive edges and make $G$ over-noisy.
Conversely, a smaller $k$ value may fail to capture the clusters' actual neighborhood structure.
Different from the previous $k$-select algorithm based on one-dimensional structural entropy maximization~\cite{li2016structural}, we consider the impact of data scale and normalize one-dimensional structural entropy as follows:
\begin{equation}
    \label{eq:correct one-dimensional structural entropy}
    H^{1}_{norm.}(G) = \frac{H^{1}(G)}{k\times |V|},
\end{equation}
where $|V|$ is the size of $V$.
$H^1_{norm.}(G)$ mitigates the impact of data scale and $k$, helping improve the model performance, as demonstrated in Sec~\ref{subsubsec:normalization}.
Then, we iterate over $k$ from 1 to $|V|$, and construct corresponding $k$-NN graph $G_k$ as well as calculating $H_{norm^1}(G_k)$.
The optimal $k$ is determined by minimum $H^1_{norm.}(G)$ in the $stable\_points$ that satisfying $H^{1}_{norm.}(G_{k}) < H^{1}_{norm.}(G_{k-1})$ and $H^{1}_{norm.}(G_{k}) < H^{1}_{norm.}(G_{k+1})$. 
The detailed process is depicted in Algorithm~\ref{alg:k selector}.
Finally, we obtain the optimal $k$-NN graph $G_{k}$ as the structured graph $G$.

\begin{algorithm}
    \caption{The structured graph construction algorithm}
    \label{alg:k selector}
    \SetAlgoRefName{1}
    \SetAlgoVlined  
    \KwIn{Data vertices set $V$}
    \KwOut{Structured graph $G$}
    $stable\_point \gets$ empty array list [];\\
    \For{$k = 1, 2, 3, \cdots , |V|$}{
        $G_k \gets$ construct initial $k$-NN graph via Eq.~\ref{eq:edge weight};\\
        $H^{1}(G_k)\gets$ calculate the one-dimensional structural entropy via Eq.~\ref{eq:1-dim se};\\
        $ H^{1}_{norm.}(G)[k]\gets$ correct $H^{1}(G_k)$ via Eq.~\ref{eq:correct one-dimensional structural entropy};\\
        \If{$H^{1}_{norm.}(G)[k-1] < H^{1}_{norm.}(G)[k-2]$ and $H^{1}_{norm.}(G)[k-1] < H^{1}_{norm.}(G)[k]$}{
            $stable\_point$ append $k-1$;
        }
    }
    \For{$k\in stable\_points$}{
        $k^{*} \gets$ select $k$ corresponding the minimum $H^{1}_{norm.}(G)[k]$ in stable point;
    }
    $G\gets$ the $k^{*}$-NN graph $G_{k*}$;\\
    Return $G$;
\end{algorithm}

\subsubsection{Two-level encoding tree optimization} \label{subsubsec:Two-dimensional encoding tree optimization}
To attain an optimal hierarchical abstracting of the dataset and uncover its underlying density distribution, we optimize the two-level encoding tree of the graph $G$ by minimizing the two-dimensional structural entropy within the structural graph $G$. 
The optimal two-level encoding tree is represented as follows:
\begin{equation}
    T=\mathop{argmin}_{\forall T:height(T) \leq 2}H^T(G), \label{7: optimal encoding tree}
\end{equation}
where $H^T(G)$ is the structural entropy defined in Eq.~\ref{eq:tree se}.
We recursively employ two optimization operators introduced from deDoc~\cite{li2018decoding}, merging and combining, to optimize the encoding tree based on the greedy policy.
The merging operator $M(T; \alpha, \gamma)$ combines two subtrees under the same parent node into one subtree, while the combining operator $C(T; \alpha, \gamma{})$ constructs a common new parent node for two subtrees.
If the structural entropy $H^T(G)$ of the encoding tree $T$ decreases after executing $M(T; \alpha, \gamma)$ or $C(T; \alpha, \gamma{})$, we consider the operator to have executed successfully and denote as $M(T; \alpha, \gamma)\downarrow$ or $C(T; \alpha, \gamma)\downarrow$.
The optimizing process is as follows.
\begin{enumerate}[leftmargin=*]
    \item Given encoding tree $T$, set $\lambda$ as the root node, $\alpha$ and $\gamma$ as the non-root nodes.
    \item If there exist $\alpha$ and $\gamma$ $\in T$ that satisfy $M(T; \alpha, \gamma)\downarrow$, execute $M(T; \alpha, \gamma)$ and obtain the updated encoding tree $T_M(\alpha, \gamma)$, then return to (2).
    \item If there exist $\alpha$ and $\gamma \in T$ that satisfy $C(T; \alpha, \gamma)\downarrow$, execute $C(T; \alpha, \gamma)$ and obtain the updated encoding tree $T_C(\alpha, \gamma)$, then return to (2).
    \item If there are no $\alpha$ and $\gamma \in T$ that satisfy $M(T; \alpha, \gamma)\downarrow$ or $C(T; \alpha, \gamma)\downarrow$, output the optimal encoding tree $T$.
\end{enumerate}
Finally, the optimal two-level encoding tree $T$ is generated. 
Notably, $T$ can also be directly utilized for clustering purposes, and its effectiveness is investigated in Sec.~\ref{subsubsec:cluster based on encoding tree}.

\begin{figure*}
\centering
\includegraphics[width=0.97\textwidth]{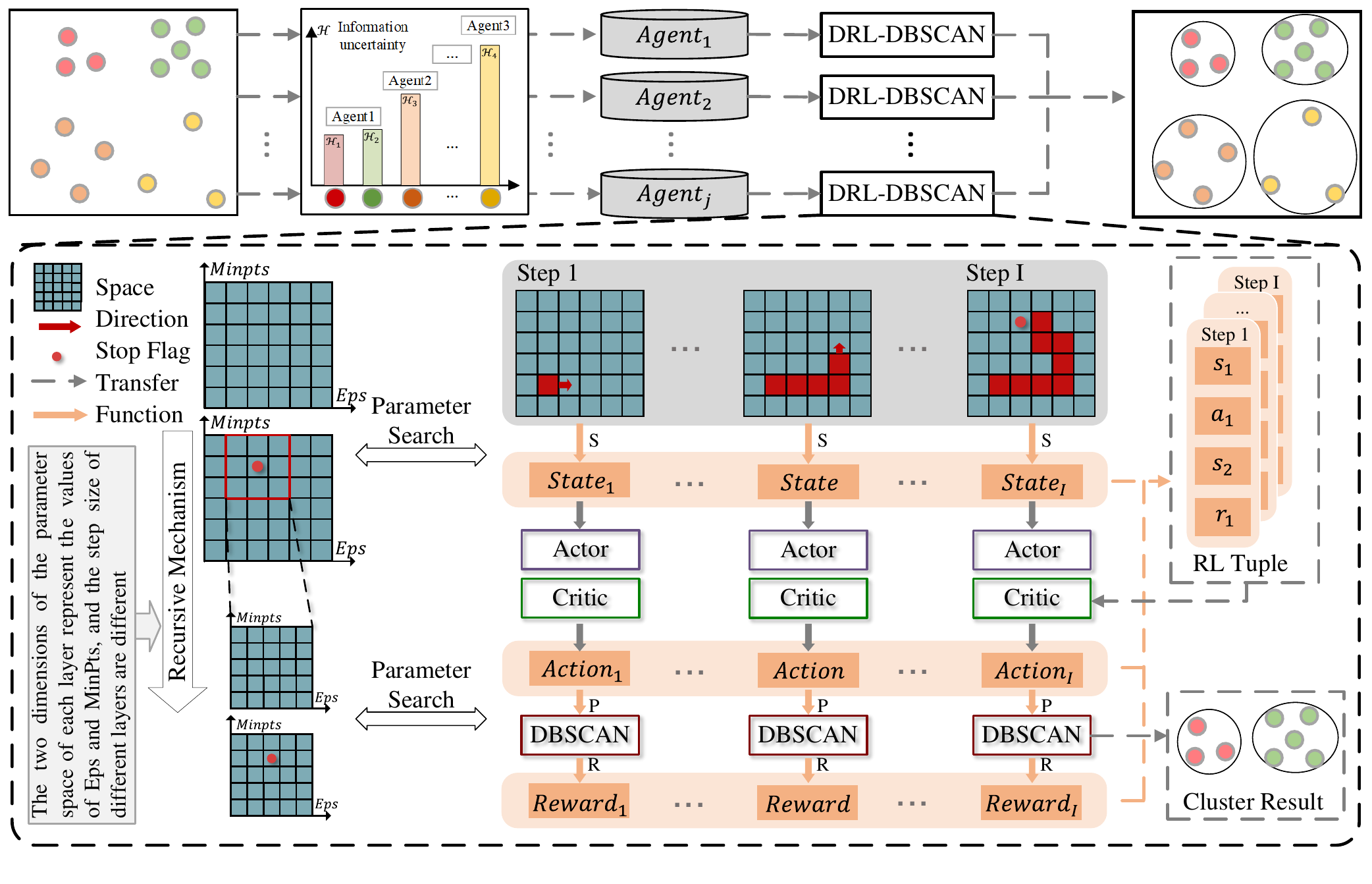}
\caption{Multi-agent deep reinforcement learning based automatic parameter search.}
\label{fig:Parameter search}
\end{figure*}

\subsubsection{Agent assignment}\label{subsubsec:agent assignment strategy}
To address the challenge of varying densities within the dataset, the main solution is to partition the initial data based on its density distribution and perform a separate agent parameter search for each partition. 
Utilizing the optimal encoding tree $T$ of the structural graph $G$, we evaluate the structural entropy of intermediate nodes using Eq.~\ref{eq:tree node se}, which provides new insights into the density differences inherent in the data.
Since the structural entropy of an intermediate node $\alpha$ is directly proportional to the number of its child nodes, we normalize it and give the formulation of information uncertainty as follows:
\begin{equation}
    \label{eq:information uncertainty}
    \mathcal{H}^{T}(G;\alpha) = \frac{H^{T}(G;\alpha)}{N_{\alpha} \times k},
\end{equation}
where $N_{\alpha}$ is the number of immediate successor nodes of $\alpha$ and $k$ is selected in Sec.~\ref{subsubsec:Structuredd graph construction}.
The normalized information uncertainty $\mathcal{H}^{T}$ displays a similar scale of values across various datasets, which is also demonstrated in Sec.~\ref{subsubsec:hyperparameter on agent allocation}. 
Hence, we simply employ the DBSCAN algorithm to cluster the intermediate nodes with similar information uncertainty into the same partition.
Specifically, as the scenario depicted in Figure~\ref{fig:allocate-agent} \emph{Allocate agents}, the intermediate nodes with information uncertainty $\mathcal{H}_1$ and $\mathcal{H}_2$ are partitioned together as their information uncertainty difference is small, and the other two intermediate nodes are partitioned separately.
This way, the leaf nodes in the encoding tree (i.e., the initial data points) are separated into different partitions. 
For each partition, we allocate a dedicated agent and conduct a parameter search specific to that agent.

\subsection{Parameter search for each agent}\label{subsec:Parameter search for each agent}
Up to this point, we have successfully determined the number of agents and partitioned the corresponding data for them. 
In this subsection, we present the fundamental Markov decision process for parameter exploration and the recursive mechanism for navigating the parameter space, as illustrated in Figure~\ref{fig:Parameter search}.

\subsubsection{Parameter search with DRL}\label{subsubsec:parameter search with DRL}
The fixed DBSCAN parameter search policy is no longer suitable for various cluster tasks. 
As a solution, we proposed DRL-DBSCAN~\cite{zhang2022automating}, a novel framework for automatic parameter search based on deep reinforcement learning (DRL). 
The core model of DRL-DBSCAN is formulated as a Markov Decision Process (MDP) with state set, action space, reward function, and policy optimization algorithm $(\mathcal{S}, \mathcal{A}, \mathcal{R}, \mathbf{P})$~\cite{mundhenk2000mdp}.

This process converts the DBSCAN parameter search into a maze game problem~\cite{zheng2012maze,bom2013pac_man} in the parameter space, with the objective of training an agent to iteratively navigate from the start point parameters to the endpoint parameters by interacting with the environment. 
The endpoint parameters obtained are considered the final search result of the game. 
The agent perceives the parameter space and DBSCAN clustering algorithm as the environment, with the search position and clustering result representing the state and the parameter adjustment direction serving as the action. 
To encourage exceptional behavior, a small number of samples are used to provide weakly supervised rewards to the agent. 
We optimize the agent's policy using the Actor-Critic \cite{konda2000actor_critic} architecture. 
Specifically, we formulate the search process for episode $e(e=1,2,\dots)$ on the corresponding partition $\mathcal{P}$ of the agent as follows:

\textit{\textbf{$\bullet$ State:}} 
To achieve an accurate and comprehensive state representation of the search environment, we construct it from global and local perspectives. 

Firstly, we describe the overall search and clustering situation by utilizing a 7-tuple to depict the global state of the $i$-th step $(i = 1, 2, \dots)$:
\begin{equation}\label{eq:global_state}
s_{global}^{(e)(i)}= P^{(e)(i)} \ {\cup}\ \mathcal{D}_{b}^{(e)(i)} \ {\cup}\  \big\{{R}_{cn}^{(e)(i)}\big\}.
\end{equation}
Here, $P^{(e)(i)}=\{Eps^{(e)(i)}, MinPts^{(e)(i)}\}$ refers to the current DBSCAN parameter. 
$\mathcal{D}_{b}^{(e)(i)}$ represents a collection of quaternary distances, comprising the distance between $Eps^{(e)(i)}$ and its corresponding space boundaries $B_{Eps,1}$ and $B_{Eps,2}$, as well as the distance between $MinPts^{(e)(i)}$ and its corresponding boundaries $B_{MinPts,1}$ and $B_{MinPts,2}$ (the specific parameter boundaries are defined in Sec.~\ref{subsubsec: Parameter Space and Recursion Mechanism}). 
The ratio ${R}_{cn}^{(e)(i)}=\frac{|\mathcal{C}^{(e)(i)}|}{|\mathcal{P}|}$ corresponds to the number of clusters $|\mathcal{C}^{(e)(i)}|$ relative to the overall number of objects in the data partition $|\mathcal{P}|$.
Secondly, to depict the state of each cluster, we define the $\{d+2\}$-tuple of the $i$-th local state of cluster $c_{n} \in \mathcal{C}$ as follows:
\begin{equation}\label{eq:local_state}
s_{local, n}^{(e)(i)}= \mathcal{X}_{center,n}^{(e)(i)} \cup  \big\{D^{(e)(i)}_{center,n},  |c_{n}^{(e)(i)}|\big\},
\end{equation}
where $\mathcal{X}_{center,n}^{(e)(i)}$ represents the feature of the central object of $c_{n}$ and $d$ is its feature dimension, $D^{(e)(i)}_{center,n}$ denotes the Euclidean distance between the cluster center object and the central object of the entire data partition, and $|c_{n}^{(e)(i)}|$ means the number of objects contained in cluster $c_{n}^{(e)(i)}$.
Thirdly, taking into account the variation in the number of clusters during different steps of the parameter search process, we use the Attention Mechanism~\cite{vaswani2017attention} to consolidate the global state and multiple local states into a state representation of fixed length:
\begin{equation}\label{eq:state}
s^{(e)(i)}= 
\sigma \Big(F_{G}(s_{global}^{(e)(i)}) \mathbin\Vert  
\sum_{c_n \in \mathcal{C}} \alpha_{att,n} \cdot 
F_{L}(s_{local, n}^{(e)(i)})\Big),
\end{equation}
where $F_{G}$ and $F_{L}$ refer to the Fully-Connected Network (FCN) for the global state and local state, respectively. 
$\sigma$ is the \texttt{ReLU} activation function, while $\mathbin\Vert$ indicates the splicing operation. 
$\alpha_{att,n}$ denotes the attention weight assigned to cluster $c_{n}$, which is formulated as follows:
\begin{equation}
\alpha_{att,n} = \frac{ \sigma \Big(
F_{S}\big(F_{G}(s_{global}^{(e)(i)}) \mathbin\Vert F_{L}(s_{local,n}^{(e)(i)})\big)\Big)}
{\sum_{c_n \in \mathcal{C}} \sigma \Big(F_{S} \big(F_{G}(s_{global}^{(e)(i)})  \mathbin\Vert  F_{L}(s_{local,n}^{(e)(i)})\big)\Big)}.
\label{equation:att}
\end{equation}
We merge the global state with the local state of each cluster separately, feed it into a fully connected network $F_{S}$ for scoring, and utilize the normalized score of each cluster as its attention weight.
This method establishes attention toward the global search environment while representing local clusters.
Additionally, it assigns distinct weights to different types of cluster information during the final state representation, thus augmenting the influence of significant clusters on the state.

\textit{\textbf{$\bullet$ Action:}} 
The action $a^{(e)(i)}$ for the episode $e$ and $i$-th step corresponds to the parameter search direction. 
To be precise, the action space $\mathcal{A}$ is defined as $\{left, right, down, up, stop\}$, where $left$ and $right$ refer to reducing or increasing the $Eps$ parameter, $down$ and $up$ correspond to reducing or increasing the $MinPts$ parameter, and $stop$ indicates the termination of the search process. 
To determine the action $a^{(e)(i)}$ based on the current state $s^{(e)(i)}$, we develop an Actor \cite{konda2000actor_critic} network as the policy network, which is described as follows:
\begin{equation}\label{eq:action}
a^{(e)(i)}= Actor(s^{(e)(i)}),
\end{equation}
where the $Actor$ is a three-layer Multi-Layer Perceptron (MLP).
The conversion process of an action from the $i$-th step to the $i+1$-th step is defined as follows:
\begin{equation}\label{eq:action2parameter}
P^{(e)(i)} \xrightarrow[]{a^{(e)(i)},\theta} P^{(e)(i+1)}.
\end{equation}
$P^{(e)(i)}$ and $P^{(e)(i+1)}$ represent the parameter combinations of ${Eps^{(e)(i)}, MinPts^{(e)(i)}}$ and ${Eps^{(e)(i+1)}, MinPts^{(e)(i+1)}}$ at the $i$-th step and the $i+1$-th step, respectively.
$\theta$ denotes the parameter adjustment step size, which will be discussed in detail in Sec.~\ref{subsubsec: Parameter Space and Recursion Mechanism}. 
It should be noted that if an action causes a parameter to go beyond its boundary, the parameter is set to the boundary value, and the corresponding boundary distance is set to $-1$ in the next step.

\textit{\textbf{$\bullet$ Reward:}}
Rewards play a crucial role in incentivizing the agent to improve its parameter search policy.
Thus, we use a small number of external metric samples as the basis for providing rewards to the agent. 
The immediate reward function for the $i$-th step is defined as follows:
\begin{equation}\label{eq:reward_function}
\mathcal{R}(s^{(e)(i)},a^{(e)(i)}) = \mathrm{NMI}\big(\mathrm{DBSCAN}(\mathcal{X},P^{(e)(i+1)})',\mathcal{Y}'\big),
\end{equation}
where $\mathrm{NMI}(,)$ represents the external metric function, Normalized Mutual Information (NMI)~\cite{estevez2009nmi} of DBSCAN clustering.
$\mathcal{X}$ denotes the feature set, and $\mathrm{DBSCAN}(\mathcal{X},P^{(e)(i+1)})'$ denotes the predicted labels for the partially labeled data.
$\mathcal{Y}'$ refers to a set of partial true labels for the data partition.
It is important to note that the labels are used solely for training purposes, and the testing process involves searching for optimal parameters on unlabeled data.

Considering the optimal parameter search action sequence for one episode involves adjusting the parameters towards the optimal parameters and stopping the search at the optimal parameters, we propose using both the maximum immediate reward for subsequent steps and the endpoint immediate reward as the reward for the $i$-th step:
\begin{equation}\label{eq:reward}
\begin{aligned}
    r^{(e)(i)}=&\beta \cdot \max \big\{\mathcal{R}(s^{(e)(m)},a^{(e)(m)})\big\}|_{m=i}^{I}\\
&+\delta \cdot \mathcal{R}(s^{(e)(I)},a^{(e)(I)}), 
\end{aligned}
\end{equation}
where $\mathcal{R}(s^{(e)(I)},a^{(e)(I)})$ is the immediate reward for the $I$-th step end point of episode $e$.
$\max$ computes the future maximum immediate reward prior to stopping the search in the current episode $e$. 
$\beta$ and $\delta$ denote the reward impact factors, where $\beta=1-\delta$.

\textit{\textbf{$\bullet$ Termination:}}
To enhance the efficiency of the search process, we design three termination conditions for a complete episode search process, including parameter search going beyond the boundaries, the search step exceeding the maximum step limit in an episode, and the action turning to $stop$. 
The formulation for these conditions is as follows:
\begin{equation}\label{eq:termination}
\left\{
\begin{array}{ll}
    \min(\mathcal{D}_{b}^{(e)(i)})<0,      &      \text{Out of bounds stop,}\\
    i>=I_{max},    &     \text{Timeout stop,}\\
    a^{(e)(i)}=stop, where \ i \ge 2,     &    \text{Action stop,}\\
\end{array} 
\right.
\end{equation}
where $I_{max}$ is the maximum step limit in an episode.


\textit{\textbf{$\bullet$ Optimization:}}
To enhance the performance of the policy network $Actor$, it is necessary to extract specific components from the network and enable the $Critc$ network to evaluate and adjust the $Actor$ network accordingly.
The parameter search process for episode $e$ is described as follows:
$1)$ Observe the current state $s^{(e)(i)}$ of DBSCAN clustering.
$2)$ Predict the action of the parameter adjustment direction, $a^{(e)(i)}$, based on $s^{(e)(i)}$ through the $Actor$ network.
$3)$ Obtain the new state, $s^{(e)(i+1)}$.
$4)$ Repeat the above steps until the end of the episode, and get a reward $r^{(e)(i)}$ for each step.
Therefore, the core element of the $i$-th step is extracted as a \emph{RL Tuple}:
\begin{equation}\label{eq:core_element}
\mathcal{T}^{(e)(i)}=(s^{(e)(i)},a^{(e)(i)},s^{(e)(i+1)},r^{(e)(i)}).
\end{equation}
We store $\mathcal{T}$ of each step in the memory buffer and sample $M$ core elements to optimize the policy network $Actor$. 
Specifically, we define a three-layer MLP as the $Critic$ to learn the action value of the state \cite{konda2000actor_critic} and use the $Critic$ network to optimize the $Actor$. 
The loss function for $Actor$ and $Critic$ is defined as follows:
\begin{equation}\label{eq:loss_critic}
\begin{aligned}
    \mathcal{L}_{c} = &{\sum}_{\mathcal{T} \in buffer}^{M} \big(r^{(e)(i)}+\gamma \cdot Critic(s^{(e)(i+1)},a^{(e)(i+1)})\\
    &-Critic(s^{(e)(i)},a^{(e)(i)})\big)^{2},
\end{aligned}
\end{equation}
\begin{equation}\label{eq:loss_actor}
    \mathcal{L}_{a} = -\frac{{\sum}_{\mathcal{T} \in buffer}^{M} Critic\big(s^{(e)(i)},Actor(s^{(e)(i)})\big)}{M}. 
\end{equation}
Here, $\gamma$ denotes the reward decay factor. 
It should be noted that we adopt the Twin Delayed Deep Deterministic Policy Gradient algorithm (TD3)~\cite{scott2018td3} for policy optimization in our framework, which is substituted with other Deep Reinforcement Learning policy optimization algorithms~\cite{lillicrap2015ddpg, konda2000actor_critic}.

\subsubsection{Parameter Space and Recursion Mechanism}\label{subsubsec: Parameter Space and Recursion Mechanism}
Here, we define the parameter space of the agent proposed in Sec.~\ref{subsubsec:parameter search with DRL} and the recursive search mechanism that starts with a broad search space with coarse-grained precision and gradually narrows it down to a smaller search space with fine-grained precision. 
Firstly, to address the fluctuation of the parameter range caused by different data distributions, we normalize the data features by transforming the maximum search range of the $Eps$ parameter into the $(0,\sqrt{d}]$ range, where $d$ is the dimension of the data features. 
As $MinPts$ must be an integer greater than 0, we propose to define a coarse-grained maximum search range for $MinPts$ based on the size of the dataset.
Secondly, in order to balance search efficiency and search precision, we propose a recursive mechanism for progressive search that narrows down the search range and increases the search precision layer by layer.
Once the agent in the previous layer reaches termination, we replace it with a new agent in the next layer and search for the optimal parameter combination $P_{o}^{(l)}=\{Eps_{o}^{(l)}, MinPts_{o}^{(l)}\}$ according to the search precision and range requirements of the corresponding layer.

The minimum and maximum search boundaries $B_{p,1}^{(l)}$ and $B_{p,2}^{(l)}$ for parameter $p \in \{Eps, MinPts\}$ in layer $l$ $(l=0,1,2,\cdots)$ are defined as follows:
\begin{equation}\label{eq:boundary}
\begin{split}
& B_{p, 1}^{(0)}: {B_{Eps, 1}^{0}=0, B_{MinPts, 1}^{0}=1},\\
& B_{p, 2}^{(0)}: {B_{Eps, 2}=\sqrt{d}, B_{MinPts, 2}=|\mathcal{P}|},\\
& B_{p,1}^{(l)} : \max \big\{ B_{p,1}^{(0)}, \ p_{o}^{(l-1)} - \frac{\pi_{p}}{2} \cdot \theta_{p}^{(l)} \big\},\\
& B_{p,2}^{(l)} : \min \big\{ p_{o}^{(l-1)} + \frac{\pi_{p}}{2} \cdot \theta_{p}^{(l)}, B_{p,2}^{(0)} \big\}, \\
\end{split}
\end{equation}
where $d$ represents the dimensionality of the data features, $|\mathcal{P}|$ is the size of the data partition allocated to the agent in Sec.~\ref{subsec:Structural entropy based agent allocation}, $B_{p,1}^{(0)}$ and $B_{p,2}^{(0)}$ are the parameter boundaries of the $0$-th layer that determine the maximum range for parameter search. 
$\pi_{p}$ represents the number of parameters that can be searched in the parameter space of parameter $p$ in each layer. $p_{o}^{(l-1)} \in P_{o}^{(l-1)}$ denotes the optimal parameter searched by the previous layer. 
For the $0$-th layer, we define $p_{o}^{(0)} \in P_{o}^{(0)}$ as the midpoint between $B_{p,1}^{(0)}$ and $B_{p,2}^{(0)}$.
Moreover, $\theta_{p}^{(l)}$ is the search step size that controls the search precision of parameter $p$ in the $l$-th layer and is defined as follows:
\begin{equation}\label{eq:theta}
\theta_{p}^{(l)} =
\left\{
\begin{array}{ll}
    \frac{\theta_{p}^{(l-1)}}{\pi_{p}},      & {if \  p \ is \ Eps;}\\
    \max \big\{\lfloor \frac{\theta_{p}^{(l-1)}}{\pi_{p}} + \frac{1}{2} \rfloor, \ 1 \big\},     & {otherwise.}\\
\end{array} 
\right.
\end{equation}
Here, $\theta_{p}^{l-1}$ represents the search step size of the previous layer, while $\theta_{p}^{0}$ is a manually defined value. The symbol $\lfloor\ \rfloor$ denotes rounding down to the nearest integer.


\subsection{Time complexity of \model{}}
\label{subsec:time complexity}
The time complexity of the agent allocation is $O(n^2+n\log^2{n})$, where $n$ is the number of nodes. 
Separately, the time complexity of constructing the structured graph is $O(n^2)$, and the time complexity of two-level encoding tree optimization is $O(n\log^2{n})$. 
As for the time complexity of parameter search, according to the recursive mechanism, the time complexity of parameter search for each agent is $O(\log{N})$, where $N$ denotes the size of the parameter space $\theta^{(0)}_p / \theta^{L}_p = (\pi_p)^L$. 
Consequently, the total time complexity of \model{} is $O(n^2+n\log^2{n}) + O(\log{N})$.

\section{Experimental Setup}\label{sec:setup}
In this section, we present our experimental platform, the chosen baselines, datasets, evaluation metrics, and the details of the experimental implementation.

\subsection{Hardware and Software}\label{subsec:hardware and software}
All experiments are performed on a Linux server consisting of an 18-core Intel i9-10980XE CPU, 256GB of RAM, and an NVIDIA RTX A6000 GPU. 
We implement all models using Pytorch 2.0 and Python 3.10. 
As for baselines, we utilize open-source implementations from the library Hyperopt\footnote{\url{http://github.com/hyperopt/hyperopt}} and Scikit-opt\footnote{\url{https://github.com/guofei9987/scikit-opt}}, or codes provided by the authors. 

\subsection{Baselines}\label{subsec:baselines}
We compare \model{} to traditional hyperparameter search schemes, meta-heuristic optimization algorithms, and existing DBSCAN parameter search methods.
As our work focuses on the automatic DBSCAN method, some other clustering algorithms based on the structural information principle~\cite{huang2024sec} or other variants of DBSCAN that require manual determination of hyperparameters for different datasets~\cite{diao2018improved, QIAN2024127329} are not selected as the baselines.
The baselines are categorized into three types:
(1) traditional parameter search methods: Rand~\cite{bergstra2012random}, BO-TPE~\cite{bergstra2011algorithms};
(2) meta-heuristic optimization algorithms: Anneal~\cite{kirkpatrick1983optimization}, PSO~\cite{shi1998parameter}, GA~\cite{lessmann2005optimizing}, DE~\cite{qin2008differential};
(3) specific DBSCAN parameter search methods: KDist (VDBSCAN)~\cite{liu2007vdbscan}, BDE (BDE-DBSCAN)~\cite{karami2014Choosing}, AMD-DBSCAN~\cite{wang2022amd}, DRL-DBSCAN~\cite{zhang2022automating}.
The details of these baselines are as follows.

\noindent
\textbf{Rand}~\cite{bergstra2012random}. 
The Rand algorithm finds better models by effectively searching a larger, less promising configuration space. 
It searches over the same domain to find models that are as good or better within a small fraction of the computation time.

\noindent
\textbf{BO-TPE}~\cite{bergstra2011algorithms}. 
The BO-TPE algorithm uses random search and two new greedy sequential methods based on the expected improvement criterion to optimize hyper-parameters. 

\noindent
\textbf{Anneal}~\cite{kirkpatrick1983optimization}. 
The Anneal introduces the idea of annealing into the field of combinatorial optimization. 
It is a stochastic optimization algorithm based on the Monte-Carlo iterative solution strategy, which is based on the similarity between the annealing process of solid matter in physics and general combinatorial optimization problems.

\noindent
\textbf{PSO}~\cite{shi1998parameter}.
The PSO algorithm is a swarm optimization algorithm in which each member continuously changes its search pattern by learning from its own experience and the experience of other members.

\noindent
\textbf{GA}~\cite{lessmann2005optimizing}.
The GA algorithm combines the genetic algorithm and support vector machine method. 
Support vector machines are used to solve classification tasks, while genetic algorithms are optimization heuristics that combine direct and random searches within a solution space.

\noindent
\textbf{DE}~\cite{qin2008differential}.
The Differential Evolution (DE) algorithm is based on evolutionary ideas such as genetic algorithms.
It is essentially a multi-objective (continuous variable) optimization algorithm for solving the overall optimal solution in multidimensional space.

\noindent
\textbf{KDist} (VDBSCAN)~\cite{liu2007vdbscan}.
The VDBSCAN algorithm selects several values of parameter $Eps$ for different densities according to a k-dist plot before adopting the traditional DBSCAN algorithm.
Clusters with varied densities simultaneity are found with different values of $Eps$.

\noindent
\textbf{BDE}~\cite{karami2014Choosing}.
BDE-DBSCAN adopts the binary coding scheme and represents each individual as a bit string. 
It leverages the Binary Differential Evolution and Tournament Selection method to simultaneously quickly and automatically specify appropriate parameter values in DBSCAN.

\noindent
\textbf{AMD-DBSCAN}~\cite{wang2022amd}
AMD-DBSCAN is an adaptive multi-density DBSCAN algorithm.
An improved parameter adaptation method for utilizing the k-dist plot is proposed to search for candidate cluster parameters in DBSCAN for each density cluster.

\noindent
\textbf{DRL-DBSCAN}~\cite{zhang2022automating} 
DRL-DBSCAN transfers the parameter search process into a Markov Decision Process and utilizes deep reinforcement learning to search for the optimal parameter for DBSCAN.

\subsection{Datesets}\label{subsec:datasets}
\begin{table}[t]
    \aboverulesep=0ex
    \belowrulesep=0ex
    \caption{Statistics of benchmark datasets.}
    \centering
    \begin{tabular}{c|c|ccc|c}
        \toprule
        Type & Dataset & Classes & Size & Dim. & Time\\
        \hline
        \multirow{9}{*}{Offline} & Aggregation & 7 & 788 & 2 & \XSolidBrush \\
        & Compound & 6 & 399 & 2 & \XSolidBrush \\
        & Pathbased & 3 & 300 & 2 & \XSolidBrush \\
        & D31 & 31 & 3100 & 2 & \XSolidBrush \\
        & Unbalance2 & 8 & 6500 & 2 & \XSolidBrush \\
        & Asymmetric & 5 & 1000 & 2 & \XSolidBrush \\
        & Skewed & 6 & 1000 & 2 & \XSolidBrush \\
        & Cure-t1-2000n-2D & 6 & 2000 & 2 & \XSolidBrush \\
        & Cure-t2-4k-2D & 7 & 4000 & 2 & \XSolidBrush \\
        \hline
        Online & Powersupply & 24 & 29928 & 2 & \Checkmark\\
        \bottomrule
        
    \end{tabular}
    \label{tab:dataset statistics}
\end{table}
We evaluate \model{} on nine artificial 2\textit{D} shape clustering benchmark datasets, including Aggregation~\cite{clustering_aggregation}, Compound~\cite{compound}, Pathbased~\cite{pathbased}, D31~\cite{d31}, Unbalance2~\cite{unbalance2_asymmetric_skewed}, Asymmetric~\cite{unbalance2_asymmetric_skewed}, Skewed~\cite{unbalance2_asymmetric_skewed}, Cure-t1-2000n-2D~\cite{cure_dataset}, Cure-t2-4k-2D~\cite{cure_dataset}, and one publicly available real-world streaming dataset, Powersupply~\cite{zhu2010stream_dataset}. 
Table~\ref{tab:dataset statistics} gives the statistics of these datasets, and further information is provided below.

\mypara{Aggregation}~\cite{clustering_aggregation}.
The Aggregation dataset is a well-known benchmark dataset for clustering algorithms. 
Its intuitive clustering structure allows for the assessment and comparison of the effectiveness of different clustering algorithms. 
Despite its apparent clustering, this dataset presents certain challenging factors, including narrow bridges between clusters and clusters of uneven sizes.

\mypara{Compound}~\cite{compound}.
The Compound dataset is a combination of three types of clusters, each exhibiting distinct characteristics. 
These include naturally distinct two clusters, two clusters separated by a small neck, and two clusters where the dense cluster is surrounded by a sparse one.

\mypara{Pathbased}~\cite{pathbased}. 
The Pathbased dataset comprises a circular cluster that has an opening near the bottom, as well as two clusters that are distributed in a Gaussian manner and are located within the circular cluster. 
Each cluster contains 100 data points.

\mypara{D31}~\cite{d31}.
The D31 dataset consists of 31 clusters, each containing 100 data points. 
The clusters are randomly placed and follow a Gaussian distribution in two dimensions.

\mypara{Unbalance2}~\cite{unbalance2_asymmetric_skewed}. 
The Unbalance2 dataset comprises three large and dense clusters, each with a size of 2000, and five small and sparse clusters, each with a size of 100. 
The proximity of the five small clusters to the large cluster poses a challenge for the clustering process.

\mypara{Asymmetric}~\cite{unbalance2_asymmetric_skewed}.
The Asymmetric dataset consists of five clusters arranged in an asymmetric structure. 
Specifically, two of the clusters are grouped together, while the remaining three clusters are separate. 
This structure poses a challenge for clustering algorithms that seek to accurately group data points into their respective clusters, especially for those with uneven sizes and densities.

\mypara{Skewed}~\cite{unbalance2_asymmetric_skewed}.
The Skewed dataset comprises six oblong clusters elongated in the 45-degree direction, which makes the data points not evenly distributed throughout the dataset.

\mypara{Cure-t1-2000n-2D}~\cite{cure_dataset}. 
The Cure-t1-2000n-2D dataset comprises several distinct clusters, including two high-density circular clusters, one large and sparse cluster, and two elliptical clusters. 
Additionally, there are narrow bridges connecting the two elliptical clusters. 
This dataset presents a significant challenge for clustering algorithms due to the presence of complex and irregularly shaped clusters, as well as the narrow bridges between some of the clusters. 

\mypara{Cure-t2-4k-2D}~\cite{cure_dataset}. 
The Cure-t2-4k-2D dataset is an extension of the Cure-t1-2000n-2D dataset, with the addition of 2000 points introduced as noise data. 
This addition significantly increases the difficulty of clustering the dataset, as the noise data further complicates the identification of the underlying clusters.

\mypara{Powersupply}~\cite{zhu2010stream_dataset}. 
The Powersupply dataset is collected from an Italy electricity company with 24 clusters and about 30,000 data points. 
This stream contains three-year power supply records from 1995 to 1998, and the task is to predict which hour (1 out of 24 hours) the current power supply belongs to.

\subsection{Evaluation metrics}\label{subsec:evaluation metrics}
We evaluate experiments based on two metrics: accuracy and efficiency. 
Specifically, we use the normalized mutual information~\cite{estevez2009nmi} (NMI) and the adjusted rand index~\cite{vinh2010ari} (ARI) to measure accuracy. 
For efficiency, we employ the DBSCAN clustering round to measure the efficiency of various models. 
Since our approach, \model{}, involves multiple agents searching for optimal parameters on separate subsets of the dataset, we align all agents' clustering results and aggregate their results for each round to obtain the final result for that round. 
In cases where one agent has fewer total rounds than others due to early stopping, we repeat the last round's result for alignment purposes.

\subsection{Implementation Details}\label{subsec:implementation details}
All experiments comprise two parts: the offline evaluation and the online evaluation. 
Due to the randomness of most algorithms, we perform ten runs on all datasets with different seeds and report the mean and variance as results, except for KDist and AMD-DBSCAN, because they are heuristic and do not involve random problems.
The structured graph construction is conducted only once for each dataset, and the selected $k$ for $k$-NN is reported in~\ref{sec:selected k}.
For the allocation of agents in \model{}, we set $Eps$ to 0.3 and $MinPts$ to 1 for all datasets. 
In the parameter search part, we use a unified label training proportion of 0.2, an $Eps$ parameter space size $\pi_{Eps}$ of 5, a $MinPts$ parameter space size $\pi_{MinPts}$ of 4, a maximum number of search steps $I_{max}$ of 30, and a reward factor $\delta$ of 0.2. 
The FCN and MLP dimensions are uniformly set to $32$ and $256$, respectively, the reward decay factor $\gamma$ of Critic is set to $0.1$, and the number of samples $M$ from the buffer is set to $16$. 
We uniformly set the maximum number of episodes to 15. 
In the offline tasks, the maximum number of recursive layers $L_{max}$ is 3, and the maximum search boundary in the $0$-th layer of $MinPts$, $B_{MinPts,2}^{(0)}$, is 0.25, while in the online task, they are set to 6 and 0.0025 times the size of data block $\mathcal{V}$, respectively. 
All baselines use the same objective function (Eq.~\ref{eq:reward_function}), parameter search space, and parameter minimum step size as our models if they support the settings.

\section{Evaluation}\label{sec:evaluation}
In this section, we conduct several experiments to evaluate the performance of \model{}. 
First, we experiment with models on nine artificial offline datasets.
Then, we evaluate models through the design of an online task. 
Additionally, we conduct the ablation study to explore the impact of each component in \model{}. 
We also present a detailed case study to demonstrate the practicality and effectiveness of \model{}.

\subsection{Offline Evaluation}\label{sec:offline}
The offline evaluation is conducted on nine artificial benchmark datasets. 
Since there is no pre-training data available in offline scenarios, each parameter search process is initiated from the beginning. 
To obtain a more accurate estimate for all models, we perform all experiments using ten different random seeds. 
The Cure-t1 and Cure-t2 datasets correspond to the Cure-t1-2000n-2D and Cure-t2-4k-2D datasets introduced in Sec.~\ref {subsec:datasets}.

\subsubsection{Accuracy and Stability Analysis}\label{subsubsec:offline acc}
We perform a parameter search process on \model{}, DRL-DBSCAN, and nine other baselines in 30 clustering rounds.
We report the best NMI and ARI metrics obtained using the optimal DBSCAN parameters identified during the search process.
The results are reported in Table~\ref{tab:offline_nmi} in the format of $mean\pm variance$. 
Specifically, for the NMI metric, \model{} shows mean improvements of 30.3\%, 30.5\%, 52.0\%, 43.2\%, 41.0\%, 144.1\%, 80.0\%, 30.6\%, 10.5\% and 8.7\% across all datasets compared to the other baselines, respectively. 
For the ARI metric, \model{} shows mean improvements of 39.8\%, 35.9\%, 55.8\%, 55.3\%, 51.3\%, 175.3\%, 153.4\%, 38.6\%, 10.0\% and 10.3\% compared to the other baselines. 
These results demonstrate the superiority of our models. 
Furthermore, the decrease in variances across various datasets demonstrates the robustness of our models. 
For \model{}, variances decrease by at most 0.367 \& 0.364, 0.356 \& 0.356, 0.271 \& 0.303, 0.299 \& 0.176, 0.442 \& 0.463, 0.2 \& 0.125, 0.264 \& 0.152, 0.359 \& 0.361, 0.219 \& 0.081 on NMI and ARI for the nine datasets compared to the other baselines.
The evident benefits regarding accuracy and robustness demonstrate that, in comparison to other hyperparameter optimization baselines with the same objective function, \model{} can consistently find exceptional parameter combinations over several rounds of parameter search.
Moreover, when compared to the submodel DRL-DBSCAN, \model{} demonstrates consistent improvements in accuracy across all nine datasets, except for the Pathbased and Cure-t2 datasets, where \model{} underperforms DRL-DBSCAN with 0.041 and 0.071 on the ARI metric, respectively. 
In addition, \model{} exhibits greater stability on six of the nine datasets compared to DRL-DBSCAN. 
This indicates that allocating different density partitions to different agents can enhance the accuracy and robustness of clustering.

\begin{figure*}[htp!]
    \centering
    \includegraphics[width=0.65\textwidth]{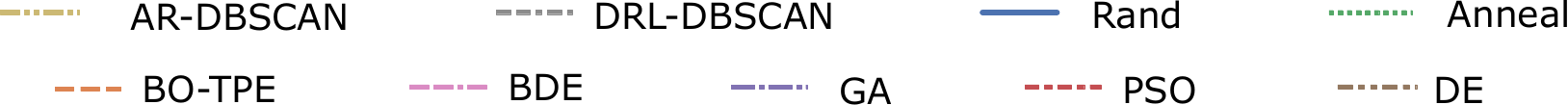}
    \vspace{-2mm}
    \subfigure[Aggregation]{
        \begin{minipage}[t]{0.4\textwidth}
            \centering
            \includegraphics[width=1\linewidth]{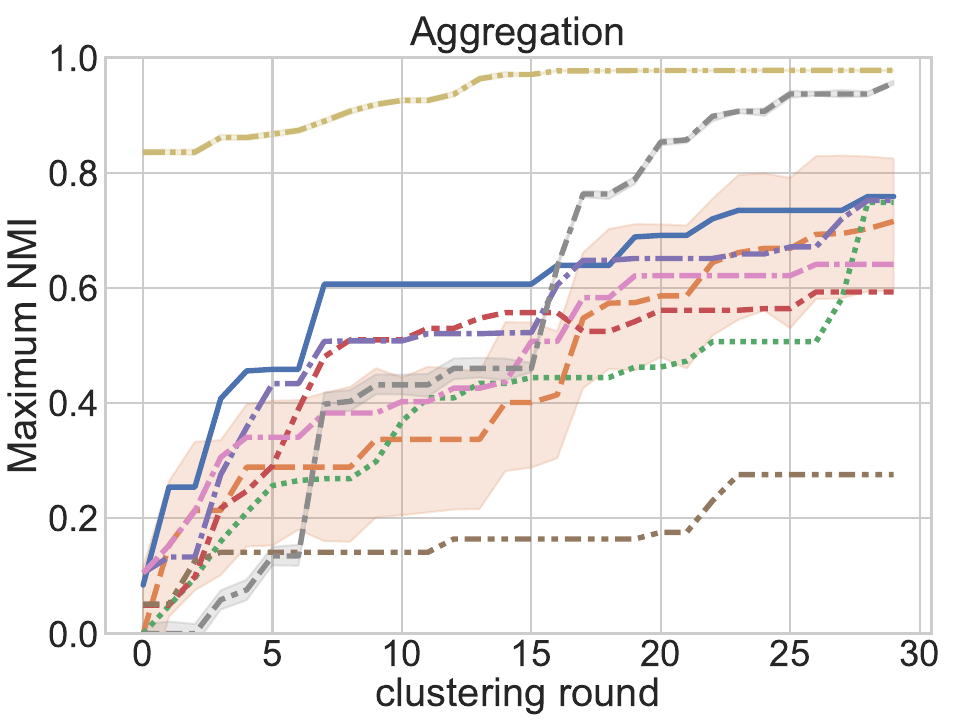}
        \end{minipage}%
    }
    \subfigure[Asymmetric]{
        \begin{minipage}[t]{0.4\textwidth}
            \centering
            \includegraphics[width=1\linewidth]{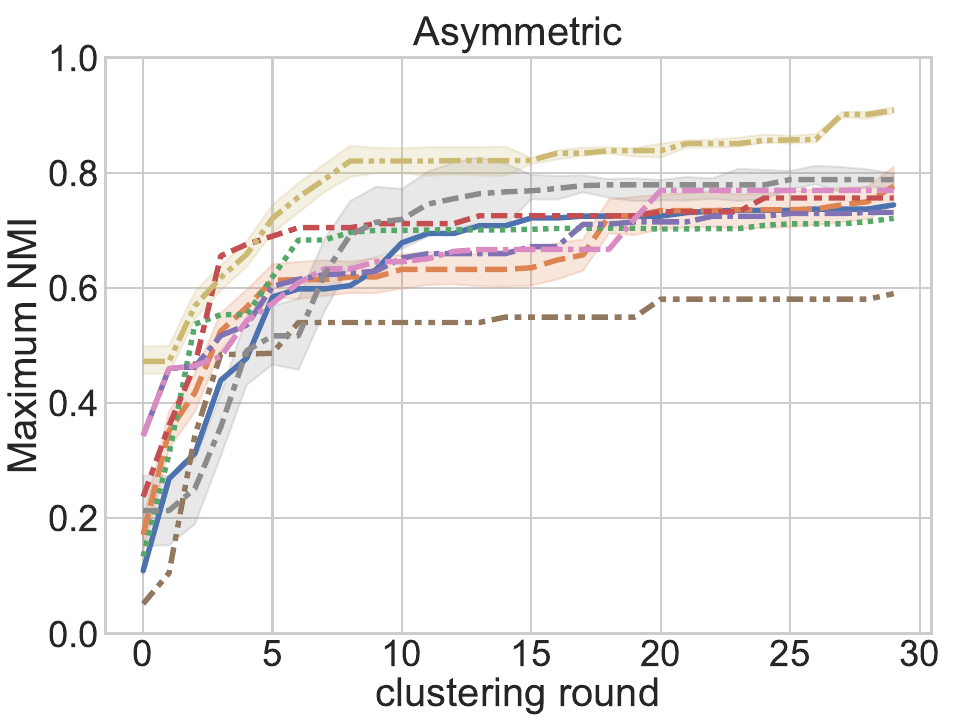}
        \end{minipage}%
    }
    \subfigure[Compound]{
        \begin{minipage}[t]{0.4\textwidth}
            \centering
            \includegraphics[width=1\linewidth]{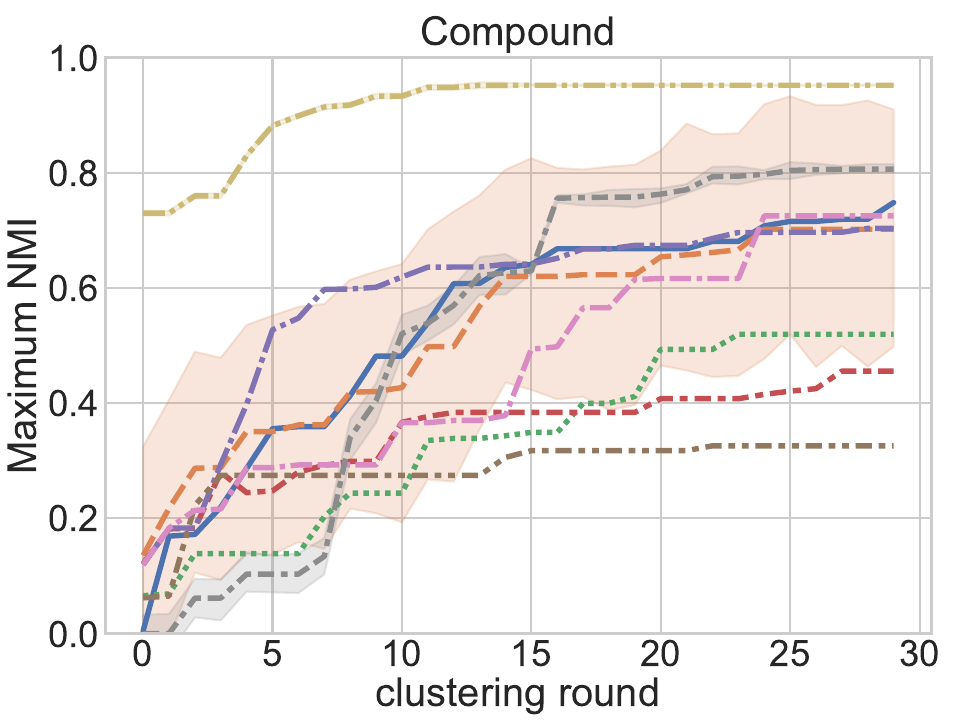}
        \end{minipage}%
    }
    \subfigure[Pathbased]{
        \begin{minipage}[t]{0.4\textwidth}
            \centering
            \includegraphics[width=1\linewidth]{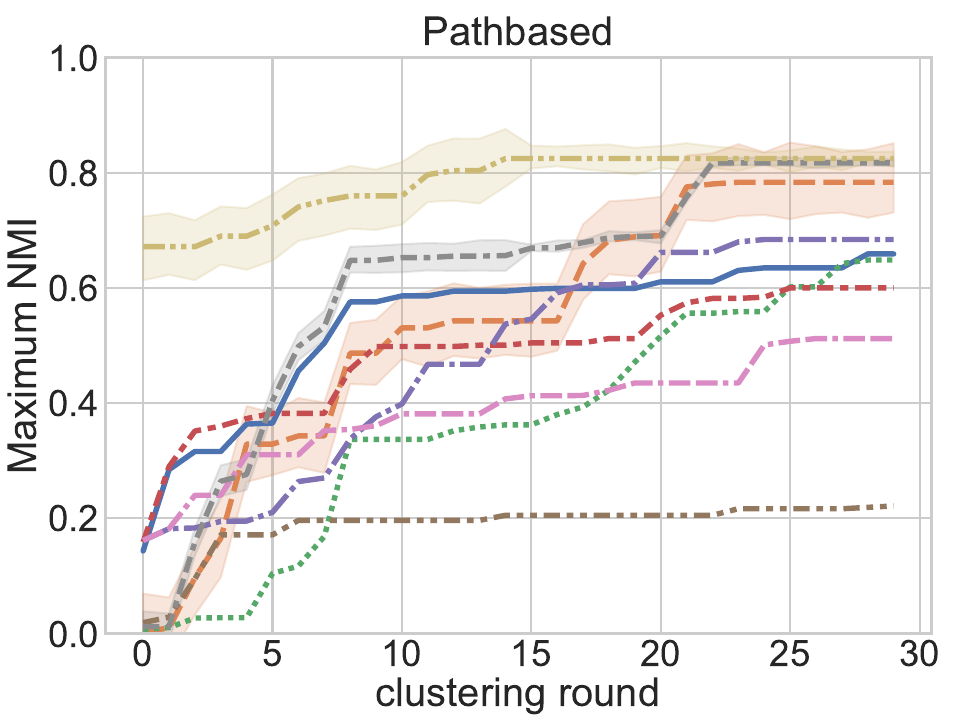}
        \end{minipage}%
    }
    \subfigure[Skewed]{
        \begin{minipage}[t]{0.4\textwidth}
            \centering
            \includegraphics[width=1\linewidth]{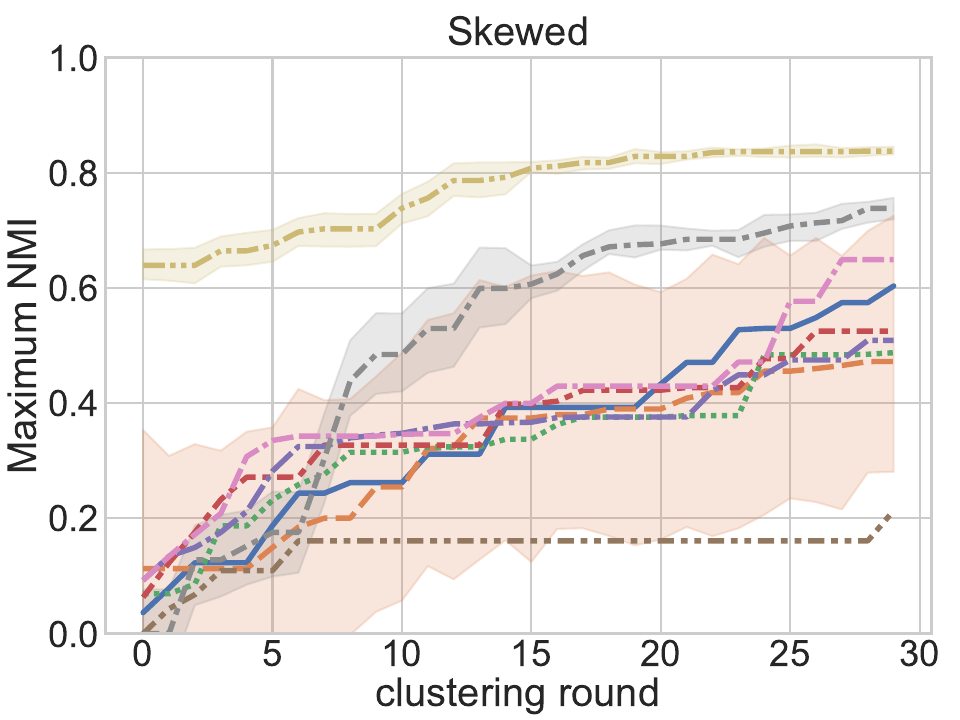}
        \end{minipage}%
    }
    \subfigure[D31]{
        \begin{minipage}[t]{0.4\textwidth}
            \centering
            \includegraphics[width=1\linewidth]{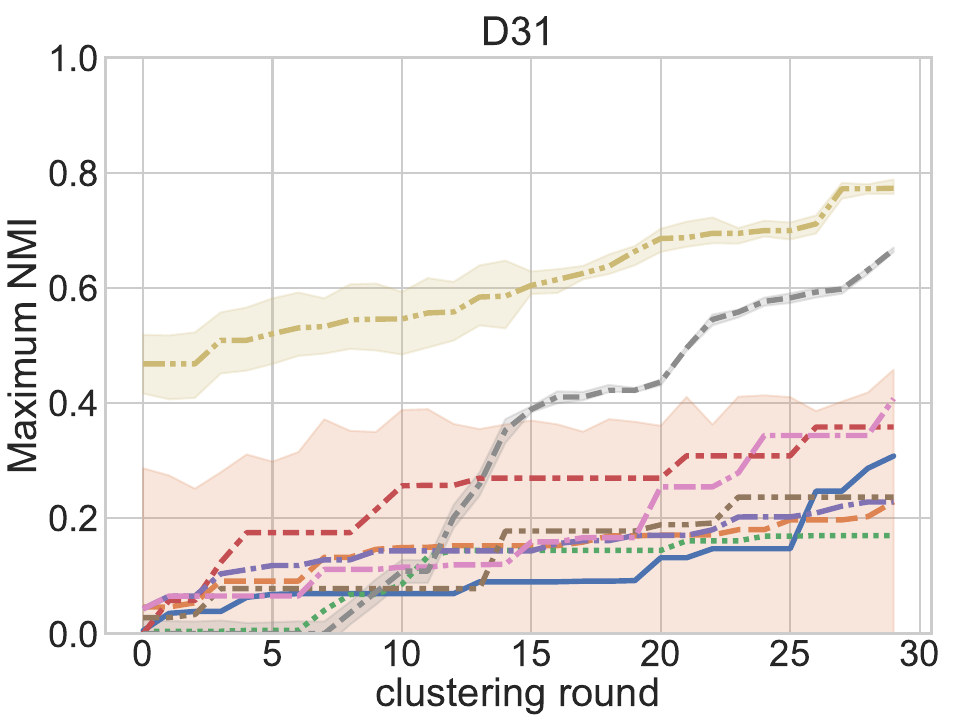}
        \end{minipage}%
    }
    \subfigure[Cure-t1]{
        \begin{minipage}[t]{0.4\textwidth}
            \centering
            \includegraphics[width=1\linewidth]{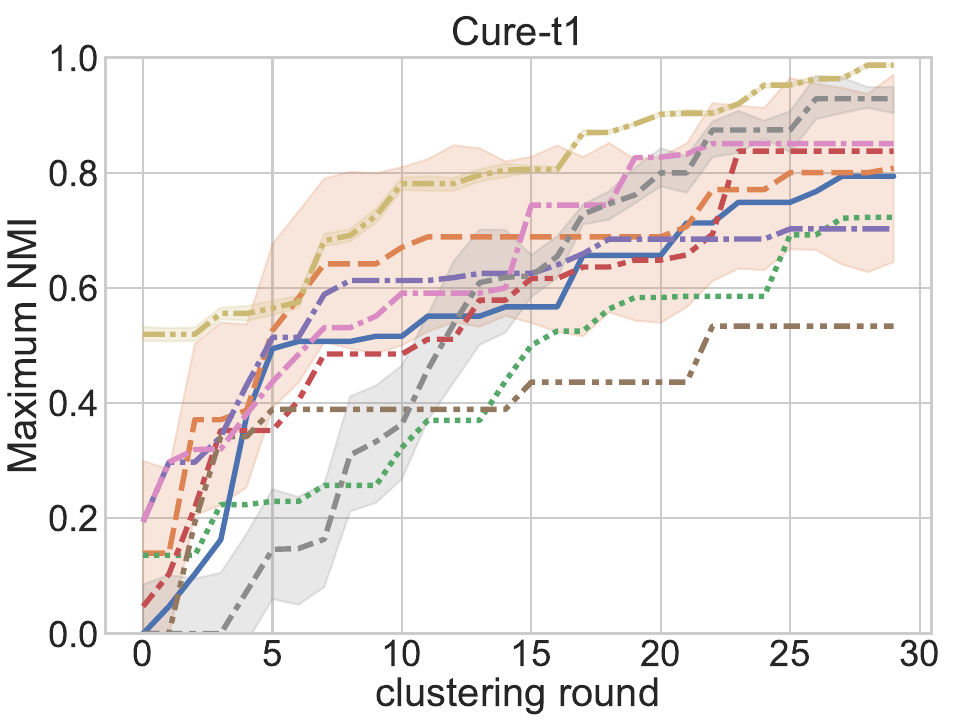}
        \end{minipage}%
    }
    \subfigure[Cure-t2]{
        \begin{minipage}[t]{0.4\textwidth}
            \centering
            \includegraphics[width=1\linewidth]{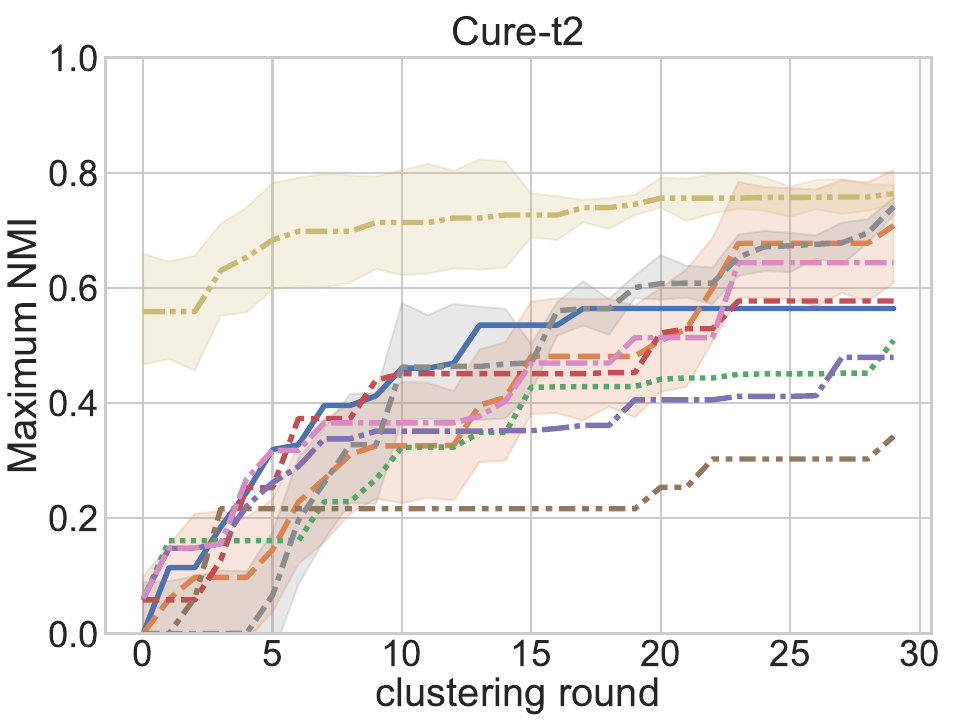}
        \end{minipage}%
    }
    \caption{Offline efficiency comparison.}
    \label{fig:offline efficiency}
\end{figure*}

\begin{sidewaystable}[htp]
    \centering
    \setlength{\tabcolsep}{0.9mm}
    \renewcommand\arraystretch{1.3}
    \aboverulesep=0ex
    \belowrulesep=0ex
    \caption{Offline evaluation of NMI and ARI performance. \textmd{The best results are bolded, and the second-best results are underlined. The Avg. column shows the mean performance of the model across all datasets while ignoring the variances.}}\label{tab:offline_nmi}
    \begin{tabular}{c|c|c|c|c|c|c|c|c|c|c|c}
       \toprule
       \textbf{Model} & \textbf{Metric} & \textbf{Aggregation} & \textbf{Compound} & \textbf{Pathbased} & \textbf{D31} & \textbf{Unbalance2} & \textbf{Asymmetric} & \textbf{Skewed} & \textbf{Cure-t1} & \textbf{Cure-t2} & \textbf{Avg.}\\
       \hline


        \multirow{2}{*}{\textbf{Rand}} & 
        NMI & .759$\pm$.114 & .748$\pm$.054 & .659$\pm$.232 & .308$\pm$.329 & .942$\pm$.053 & .744$\pm$.059 & .604$\pm$.176 & .794$\pm$.156 & .564$\pm$.266 & .684\\

        & ARI & .679$\pm$.163 & .726$\pm$.040 & .634$\pm$.211 & .138$\pm$.262 & .971$\pm$.085 & .550$\pm$.129 & .419$\pm$.187& .719$\pm$.236 & .482$\pm$.270 & .591\\
        \hline
        
        \multirow{2}{*}{\textbf{BO-TPE}} &
        NMI & .716$\pm$.141 & .702$\pm$.241 & .784$\pm$.069 & .229$\pm$.243 & .954$\pm$.005 & .777$\pm$.034 & .472$\pm$.253 & .808$\pm$.173 & .708$\pm$.115 & .683\\

        & ARI & .627$\pm$.185 & .679$\pm$.238 & .789$\pm$.104 & .044$\pm$.046 & .989$\pm$.003 & .621$\pm$.106 & .312$\pm$.186 & .771$\pm$.222 & \underline{.642$\pm$.175} & .608\\
        \hline

        \multirow{2}{*}{\textbf{Anneal}} &
        NMI & .749$\pm$.271 & .520$\pm$.361 & .649$\pm$.237 & .170$\pm$.193 & .744$\pm$.395 & .721$\pm$.190 & .487$\pm$.296 & .724$\pm$.356 & .509$\pm$.323 & .586\\

        & ARI & .704$\pm$.272 & .513$\pm$.353 & .657$\pm$.250 & .034$\pm$.042 & .770$\pm$.409 & .567$\pm$.192 & .347$\pm$.236 & .708$\pm$.360 & .473$\pm$.306 & .530\\
        \hline
        
        \multirow{2}{*}{\textbf{PSO}} &
        NMI & .593$\pm$.354 & .455$\pm$.335 & .600$\pm$.278 & .359$\pm$.334 & .893$\pm$.221 & .756$\pm$.052 & .525$\pm$.260 & .840$\pm$.200 & .578$\pm$.245 & .622\\
        & ARI & .512$\pm$.367 & .420$\pm$.361 & .552$\pm$.379 & .088$\pm$.219 & .926$\pm$.232 & .584$\pm$.117 & .374$\pm$.227 & .805$\pm$.240 & .523$\pm$.251 & .532\\
        \hline
        
        \multirow{2}{*}{\textbf{GA}} &
        NMI & .752$\pm$.146 & .704$\pm$.254 & .684$\pm$.188 & .229$\pm$.202 & .896$\pm$.193 & .731$\pm$.070 & .509$\pm$.268 & .703$\pm$.269 & .480$\pm$.285 & .632\\
        & ARI & .679$\pm$.192 & .684$\pm$.237 & .674$\pm$.260 & .040$\pm$.041 & .916$\pm$.216 & .538$\pm$.128 & .350$\pm$.220 & .639$\pm$.295 & .396$\pm$.284 & .546\\
        \hline
        
        \multirow{2}{*}{\textbf{DE}} &
        NMI & .277$\pm$.372 & .326$\pm$.354 & .222$\pm$.277 & .236$\pm$.259 & .592$\pm$.446 & .590$\pm$.228 & .214$\pm$.277 & .534$\pm$.372 & .342$\pm$.315 & .365\\
        & ARI & .253$\pm$.354 & .314$\pm$.343 & .184$\pm$.284 & .061$\pm$.087 & .606$\pm$.464 & .398$\pm$.170 & .136$\pm$.191 & .467$\pm$.377 & .283$\pm$.286 & .300\\
        \hline
        
        \multirow{2}{*}{\textbf{KDist}} &
        NMI & .601 & .392 & .404 & .074 & \underline{.979} & .468 & .516 & .449 & .569 & .495\\
        & ARI & .524 & .391 & .381 & .004 & \underline{.998} & .122 & .154 & .127 & .235 & .326\\
        \hline
        
        \multirow{2}{*}{\textbf{BDE}} &
        NMI & .634$\pm$.281 & .723$\pm$.251 & .512$\pm$.332  & .414$\pm$.361 & .945$\pm$.047 & .770$\pm$.050 & .649$\pm$.152 & .851$\pm$.146 & .644$\pm$.181 & .682\\
        & ARI & .542$\pm$.281 & .704$\pm$.251 & .482$\pm$.397  & .207$\pm$.280 & .975$\pm$.066 & .605$\pm$.130 & .475$\pm$.174 & .800$\pm$.234 & .577$\pm$.206 & .596\\
        \hline

        \multirow{2}{*}{\textbf{AMD-DBSCAN}} &
        NMI & .946 & \underline{.875} & .607  & \textbf{.886} & .941 & .762 & .718 & .722 & \textbf{.793} & .806\\
        & ARI & \underline{.979} & \underline{.886} & .573  & \textbf{.754} & .984 & .668 & .558 & .604 & \textbf{.749} & \underline{.751}\\
        \hline
        
        \multirow{2}{*}{\textbf{DRL-DBSCAN}} &
        NMI & \underline{.956$\pm$.021} & .784$\pm$.039 & \underline{.817$\pm$.028} & .667$\pm$.024 & .952$\pm$.002 & \underline{.789$\pm$.065} & \underline{.738$\pm$.079} & \underline{.929$\pm$.109} & .745$\pm$.113 & \underline{.820} \\

        & ARI & .959$\pm$.034 & .760$\pm$.032 & \textbf{.849$\pm$.038} & .262$\pm$.017 & .988$\pm$.001 & \underline{.684$\pm$.119} & \underline{.600$\pm$.134} & \underline{.931$\pm$.125} & \underline{.705$\pm$.110} & .749\\
        \hline
        
        \multirow{2}{*}{\textbf{AR-DBSCAN}} &
        NMI & \textbf{.978$\pm$.005} & \textbf{.951$\pm$.005} & \textbf{.825$\pm$.061} & \underline{.773$\pm$.062} & \textbf{.995$\pm$.004} & \textbf{.909$\pm$.028} & \textbf{.838$\pm$.032} & \textbf{.987$\pm$.013} & \underline{.764$\pm$.104} & \textbf{.891}\\

        & ARI & \textbf{.987$\pm$.003} & \textbf{.945$\pm$.005} & \underline{.808$\pm$.094} & \underline{.400$\pm$.104} & \textbf{.999$\pm$.001} & \textbf{.899$\pm$.067} & \textbf{.772$\pm$.084} & \textbf{.994$\pm$.016} & .634$\pm$.225 & \textbf{.826}\\

       \bottomrule
    \end{tabular}
\end{sidewaystable}

It is notable that DE underperforms other baselines in terms of accuracy due to its bias towards continuous parameters during parameter search, which requires more iterations to achieve optimal results. 
BO-TPE, on the other hand, uses a probabilistic surrogate model to learn from previously searched parameter combinations, effectively balancing exploration and exploitation. 
This approach provides advantages over other baselines when applied to multiple datasets. 
However, all these baselines are found to be stuck in high variances due to limited cluster rounds. 
KDist demonstrates strong performance on the Unbalance2 dataset, as the clusters within Unbalance2 exhibit significant density disparities and well-defined cluster boundaries. 
However, it underperforms in comparison to other methods when applied to the D31 dataset. 
This is due to the uniform density of clusters within the D31 dataset, which violates the underlying assumption of density variation made by the KDist algorithm.
In contrast, AMD-DBSCAN achieves the best performance in the D31 dataset, as it utilizes extra distance information to obtain a candidate $Eps$ list, which is very helpful for this dataset.
\model{} and DRL-DBSCAN use a recursive structure that progressively narrows the search space of parameters for each layer while also learning from historical experience. 
The superior performance demonstrates that this approach is more suitable for searching DBSCAN clustering parameter combinations. 
Additionally, \model{} introduces the pre-partition of different densities based on structural entropy, which further improves the cluster accuracy and robustness by easing the optimal parameter combination collision of different density clusters.

\subsubsection{Efficiency Analysis}\label{subsubsec:offline efficiency}

To showcase the parameter searching efficiency of our models, we report the average historical maximum normalized mutual information (NMI) results for \model{}, DRL-DBSCAN, and the other baseline algorithms on eight datasets, with respect to the number of clustering rounds.
The results are depicted in Figure~\ref{fig:offline efficiency}, with shaded areas indicating the range of fluctuations (variance) of NMI across multiple runs. 
For comparison, the shaded area representing the variance of NMI for the BO-TPE algorithm is also displayed as a representative of the baseline algorithms. 
It is worth noting that efficiency in this context refers to the efficiency of parameter searching.
\model{} converges within 15 rounds and achieves about two times speedup in clustering convergence rounds compared to other baselines.
Besides, \model{} maintains an advantage over all datasets at any clustering round and converges to higher accuracy.
We analyze this from two aspects: 
1) The pre-partitioning based on a two-level encoding tree demonstrates its ability to cluster and provides a sufficiently good pre-clustering result. 
2) The allocation of agents for different density partitions helps to improve the performance of the DRL-DBSCAN submodel as it increases the convergence upper bound of DRL-DBSCAN. 
We also observe that the shaded area of the curves for \model{} gradually decreases over time, while the shaded area of BO-TPE tends to remain constant. 
This improvement in stability is attributed to the use of deep reinforcement learning, which allows the action decider ($Actor$) to learn and improve gradually over time.

\subsection{Online Evaluation}\label{subsec:online evaluation}
To comprehensively explore the \model{} performance, we design an online evaluation on the real-world streaming dataset Powersupply.
The Powersupply dataset is divided into eight blocks of equal size. 
All methods are initialized before the start of each block, except for DRL-DBSCAN.
Its retaining mode DRL$re$ is re-initialized before each block, but the continuous training mode DRL$con$ continues training from the last block.

\begin{sidewaystable}[htp]
    \centering
    \aboverulesep=0ex
    \belowrulesep=0ex
    \renewcommand\arraystretch{1.5}
    \setlength{\tabcolsep}{0.4mm}
    \caption{Online evaluation. \textmd{The best results are bolded, and the second-best results are underlined.}}\label{tab:online_nmi}    
    \begin{tabular}{c|c|c|c|c|c|c|c|c|c|c|c|c|c}
        \toprule

        \textbf{Block} & \textbf{Metric} & \textbf{Rand} & \textbf{BO-TPE} & \textbf{Anneal} & \textbf{PSO} & \textbf{GA} & \textbf{DE} & \textbf{KDist} & \textbf{BDE} 
        & \textbf{AMD-DBSCAN}
        & \textbf{DRL$_{re}$} & \textbf{DRL$_{con}$} & \textbf{\model{}} \\
        \hline

        \multirow{2}{*}{\textbf{$\mathcal{V}_{1}$}} &
        NMI & .082$\pm$.072 & .152$\pm$.039 & .104$\pm$.085 & .096$\pm$.081 & .130$\pm$.069 & .024$\pm$.046 & .170 & .067$\pm$.073 & .150 &\underline{.176$\pm$.004} & .176$\pm$.005 & \textbf{.179$\pm$.007}\\
        & ARI & .021$\pm$.022 & .038$\pm$.014 & .027$\pm$.022 & .024$\pm$.021 & .032$\pm$.021 & .005$\pm$.014 & .002 & .014$\pm$.020 &
        .045
        & \underline{.047$\pm$.005} & .046$\pm$.003 & \textbf{.049$\pm$.005}\\
        \hline
        
        \multirow{2}{*}{\textbf{$\mathcal{V}_{2}$}} &
        NMI & .081$\pm$.048 & 
        .096$\pm$.043 & .048$\pm$.058 & .066$\pm$.051 & .114$\pm$.036 & .072$\pm$.050 & \textbf{.178} & .072$\pm$.050 & 
        .117
        &.135$\pm$.005 & \underline{.139$\pm$.003} & .138$\pm$.012\\
        & ARI & .020$\pm$.014 & .024$\pm$.012 & .012$\pm$.014 & .016$\pm$.014 & .027$\pm$.010 & .017$\pm$.013 & .007 & .016$\pm$.013 
        &
        .029
        & .033$\pm$.003 & \underline{.034$\pm$.002} & \textbf{.035$\pm$.002}\\
        \hline
        
        \multirow{2}{*}{\textbf{$\mathcal{V}_{3}$}} &
        NMI & .159$\pm$.085 & .148$\pm$.098 & .071$\pm$.090 & .193$\pm$.069 & .160$\pm$.102 & .131$\pm$.099 & .000 & .155$\pm$.100 
        &
        .215
        & \textbf{.238$\pm$.009} & \underline{.236$\pm$.007} & \textbf{.238$\pm$.009}\\
        & ARI & .041$\pm$.029 & .041$\pm$.028 & .018$\pm$.024 & .055$\pm$.023 & .047$\pm$.032 & .036$\pm$.029 & .000 & .044$\pm$.029 
        &
        .065
        & \textbf{.074$\pm$.007} & \underline{.071$\pm$.004} & \textbf{.074$\pm$.007}\\
        \hline
        
        \multirow{2}{*}{\textbf{$\mathcal{V}_{4}$}} &
        NMI & .087$\pm$.044 & .110$\pm$.012 & .048$\pm$.057 & .087$\pm$.045 & .077$\pm$.053 & .023$\pm$.034 & \textbf{.191} & .105$\pm$.036 
        &
        .113
        & .117$\pm$.011 & .115$\pm$.014 & \underline{.122$\pm$.003}\\
        & ARI & .020$\pm$.010 & .024$\pm$.002 & .010$\pm$.012 & .020$\pm$.010 & .016$\pm$.011 & .005$\pm$.009 & .009 & .021$\pm$.008 & .024
        & \underline{.025$\pm$.002} & \textbf{.025$\pm$.001} & .024$\pm$.002\\
        \hline
        
        \multirow{2}{*}{\textbf{$\mathcal{V}_{5}$}} &
        NMI & .107$\pm$.095 & .181$\pm$.061 & .064$\pm$.098 & .172$\pm$.086 & .148$\pm$.097 & .078$\pm$.097 & .000 & .118$\pm$.100 
        &
        .204
        & .218$\pm$.014 &  \underline{.219$\pm$.005} & \textbf{.223$\pm$.004}\\
        & ARI & .026$\pm$.027 & .044$\pm$.018 & .017$\pm$.027 & .050$\pm$.026 & .040$\pm$.011 & .019$\pm$.025 & .000 & .033$\pm$.032 
        &
        .061
        & .058$\pm$.009 & \underline{.066$\pm$.004} & \textbf{.066$\pm$.003}\\
        \hline
        
        \multirow{2}{*}{\textbf{$\mathcal{V}_{6}$}} &
        NMI & .167$\pm$.057 & .199$\pm$.018 & .130$\pm$.087 & .128$\pm$.084 & .170$\pm$.059 & .054$\pm$.083 & .104 & .156$\pm$.071 
        &
        .177
        & \textbf{.204$\pm$.003} & \underline{.203$\pm$.004} & .196$\pm$.019\\
        & ARI & .046$\pm$.019 & .051$\pm$.017 & .036$\pm$.027 & .034$\pm$.024 & .047$\pm$.019 & .015$\pm$.023 & .004 & .043$\pm$.024 
        &
        .051
        & \underline{.059$\pm$.003} & \textbf{.062$\pm$.002} & .055$\pm$.012\\
        \hline
        
        \multirow{2}{*}{\textbf{$\mathcal{V}_{7}$}} &
        NMI & .086$\pm$.036 & .130$\pm$.010 & .104$\pm$.053 & .105$\pm$.040 & .075$\pm$.056 & .039$\pm$.046 & .000 & .058$\pm$.056 
        &
        .118
        & \textbf{.133$\pm$.005} & \underline{.131$\pm$.006} & \textbf{.133$\pm$.005}\\
        & ARI & .018$\pm$.011 & .033$\pm$.003 & .026$\pm$.013 & .024$\pm$.013 & .017$\pm$.016 & .008$\pm$.012 & .000 & .013$\pm$.016 
        &
        .029
        & \textbf{.036$\pm$.001} & \underline{.034$\pm$.002} & \textbf{.036$\pm$.001}\\
        \hline
        
        \multirow{2}{*}{\textbf{$\mathcal{V}_{8}$}} &
        NMI & .129$\pm$.072 & .173$\pm$.058 & .097$\pm$.097 & .120$\pm$.086 & .156$\pm$.073 & .024$\pm$.054 & .000 & .064$\pm$.078 
        &
        .164
        & \textbf{.204$\pm$.009} & \underline{.198$\pm$.003} & \textbf{.204$\pm$.009}\\
        & ARI & .030$\pm$.023 & .049$\pm$.018 & .028$\pm$.028 & .032$\pm$.026 & .042$\pm$.023 & .005$\pm$.013 & .000 & .015$\pm$.023 
        &
        .047
        & \underline{.057$\pm$.004} & \textbf{.058$\pm$.003} & \underline{.057$\pm$.004}\\
        
        \bottomrule
        \end{tabular}
    
\end{sidewaystable}

We report the NMI and ARI results in Table~\ref{tab:online_nmi}.
Since the $stop$ action may lead to an automatic termination mechanism for the search process in \model{}, we set a maximum number of search rounds to ensure that the experimental conditions for all baselines are synchronized.
Specifically, we use the average number of clustering rounds consumed when \model{} is automatically terminated as the maximum round (30 for Table~\ref{tab:online_nmi}). 
Compared to baselines other than DRL-DBSCAN, \model{} consistently shows improvement across all eight data blocks. 
Although K-Dist can determine parameters without relying on labels and iterations and achieves the best performance in blocks $\mathcal{V}2$ and $\mathcal{V}3$ according to the NMI metric, it exhibits considerable instability and may even yield 0 in some data blocks. 
Additionally, \model{} outperforms DRL-DBSCAN in four out of eight data blocks on both the NMI and ARI metrics, performing worse only in three blocks on the ARI metric.
These results demonstrate the capability of \model{} to cluster streaming data, emphasizing the accuracy and stability benefit of our agent allocation method and the learnable DBSCAN parameter search.

\subsection{Ablation Study}\label{subsec:ablation study}
In the ablation study, we investigate the effects of normalization on edge weight $W$ and one-dimensional structural entropy $H^1(G)$, the capabilities of the optimal two-level encoding tree, the sensitivity of hyperparameters in the agent allocation and the parameter search process, as well as the role of the recursive mechanism.

\begin{table}[t]
    \aboverulesep=0ex
    \belowrulesep=0ex
    \caption{The impact of normalization. \textmd{The best results are bolded, and the second-best results are underlined. \textit{w/o norm. $W$} refers to the model without edge weight normalization, and \textit{w/o norm. $H^1(G)$} indicates the absence of one-dimensional structural entropy normalization.}}\label{tab:normalization}
    \centering
        \begin{tabular}{c|ccc}
        \toprule
        \textbf{Dataset} & \textbf{w/o norm. $W$} & \textbf{w/o norm. $H^1(G)$} & \textbf{\model{}}
        \\
        \hline
        Aggregation & \textbf{.992$\pm$.005} & .977$\pm$.005 & \underline{.978$\pm$.005} \\
        Compound& .944$\pm$.016 & \underline{.946$\pm$.008} & \textbf{.951$\pm$.005} \\
        Pathbased & .737$\pm$.228 & \underline{.783$\pm$.160} & \textbf{.825$\pm$.061} \\
        D31& \underline{.729$\pm$.022} & .692$\pm$.266 & \textbf{.773$\pm$.062} \\
        Unbalance & \underline{.995$\pm$.006} & .990$\pm$.004 & \textbf{.995$\pm$.004}\\
        Asymmetric& .907$\pm$.022 & \textbf{.910$\pm$.021} & \underline{.909$\pm$.028} \\
        Skewed& .773$\pm$.042 & \underline{.776$\pm$.044} &  \textbf{.838$\pm$.032}\\
        Cure-t1 & \underline{.985$\pm$.012} & .960$\pm$.074 & \textbf{.987$\pm$.013}\\
        Cure-t2 & \textbf{.938$\pm$.022} & .757$\pm$.102 & \underline{.764$\pm$.104}\\
        \bottomrule
        \end{tabular}
\end{table}
\subsubsection{Normalization}\label{subsubsec:normalization}
In Sec.~\ref{subsubsec:Structuredd graph construction}, we propose to normalize the edge weight in Eq.~\ref{eq:edge weight} and the one-dimensional structural entropy in Eq.~\ref{eq:correct one-dimensional structural entropy}. 
Here, we experiment with the impact of normalization on \model{} and report the NMI results in Table~\ref{tab:normalization}. 
The normalization of $W$ and $H^1(G)$ enhances the model performance on six of nine datasets. 
On the Pathbased dataset, the normalization on edge weight $W$ and one-dimensional structural entropy $H^(1)(G)$ leads to an 11.9\% and 5.4\% improvement in performance, respectively, while simultaneously decreasing the variance by 0.167 and 0.099. We also observe that not normalizing the edge weight is a preferable choice on the Cure-t2 dataset. This is attributed to the fact that half of the data points in the Cure-t2 dataset are noise.
These overall improvements in performance and stability provide evidence of the positive effect of normalization in our model.

\begin{table}[t]
    \aboverulesep=0ex
    \belowrulesep=0ex
    \caption{Comparition the clustering results of \model{}, optimal two-level encode tree, and DRL-DBSCAN. \textmd{The best results are bolded, and the second-best results are underlined. The SE column is the clustering results of the optimal two-level encode tree.}}\label{tab:cluster based on encoding tree}
    \centering
    \setlength{\tabcolsep}{3.2mm}{
        \begin{tabular}{c|ccc}
        \toprule
        \textbf{Dataset} & \textbf{SE} & \textbf{DRL-DBSCAN} & \textbf{\model{}}
        \\
        \hline
        Aggregation& .836 & \underline{.956$\pm$.021} & \textbf{.978$\pm$.005} \\
        Compound& \textbf{.952} & .784$\pm$.039 & \underline{.951$\pm$.005} \\
        Pathbased& \textbf{.913} & .817$\pm$.028 & \underline{.825$\pm$.061} \\
        D31 & .632 & \underline{.667$\pm$.024} & \textbf{.773$\pm$.062}\\
        Unbalance2 & \underline{.980} & .952$\pm$.002 & \textbf{.995$\pm$.004} \\
        Asymmetric & .737 & \underline{.789$\pm$.065} & \textbf{.909$\pm$.028} \\
        Skewed& .639 & \underline{.738$\pm$.079} & \textbf{.838$\pm$.032} \\
        Cure-t1 & .903 & \underline{.929$\pm$.109} & \textbf{.987$\pm$.013}\\
        Cure-t2& .561 & \underline{.745$\pm$.113} & \textbf{.764$\pm$.104} \\
        \bottomrule
        \end{tabular}
        }
\end{table}


\begin{figure*}[htp]
    \centering
    \subfigure[Hyperparameter sensitivity in agent allocation]{
        \begin{minipage}[t]{0.4\textwidth}
            \centering
            \includegraphics[width=1\linewidth]{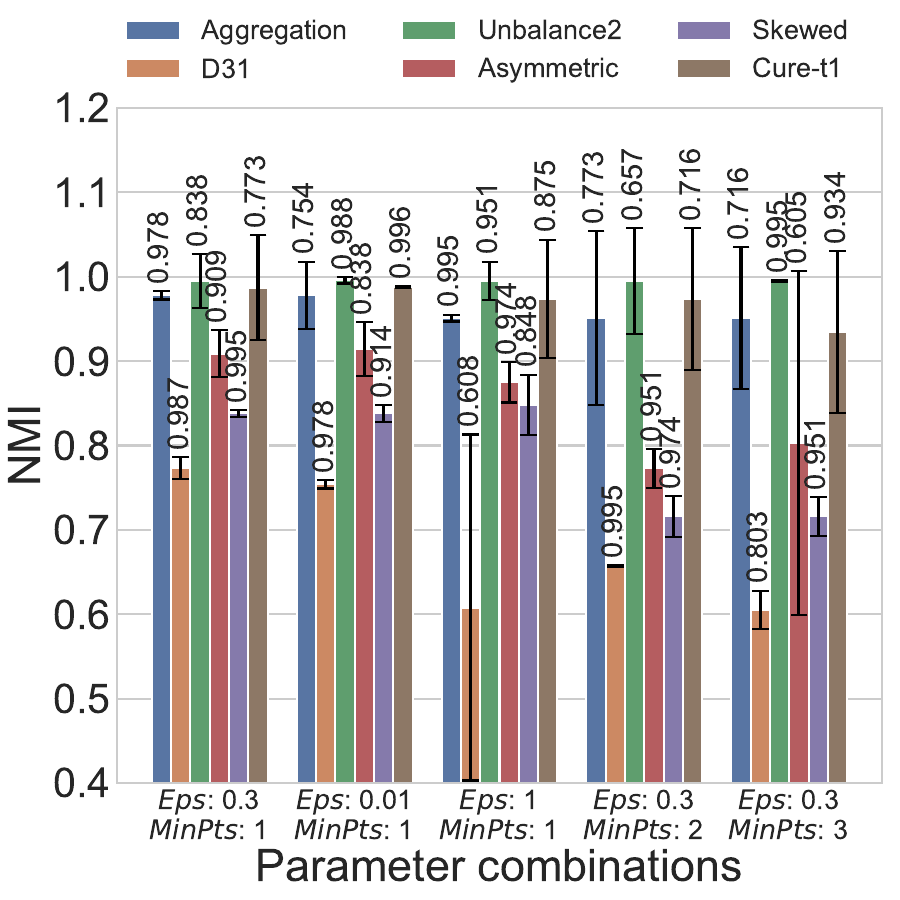}
        \end{minipage}%
    }
    \subfigure[Information uncertainty distribution]{
        \begin{minipage}[t]{0.4\textwidth}
            \centering
            \includegraphics[width=1\linewidth]{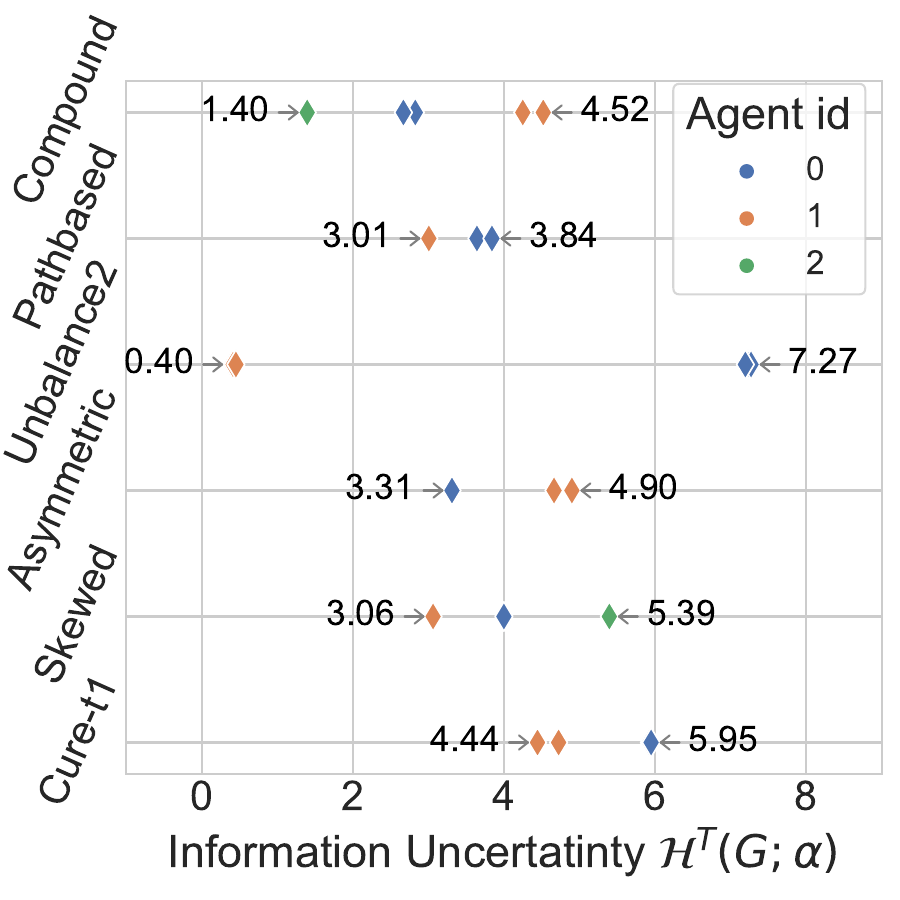}
        \end{minipage}%
    }
    \subfigure[Agent allocation and corresponding partition]{
        \begin{minipage}[t]{0.8\textwidth}
            \centering
            \includegraphics[width=1\linewidth]{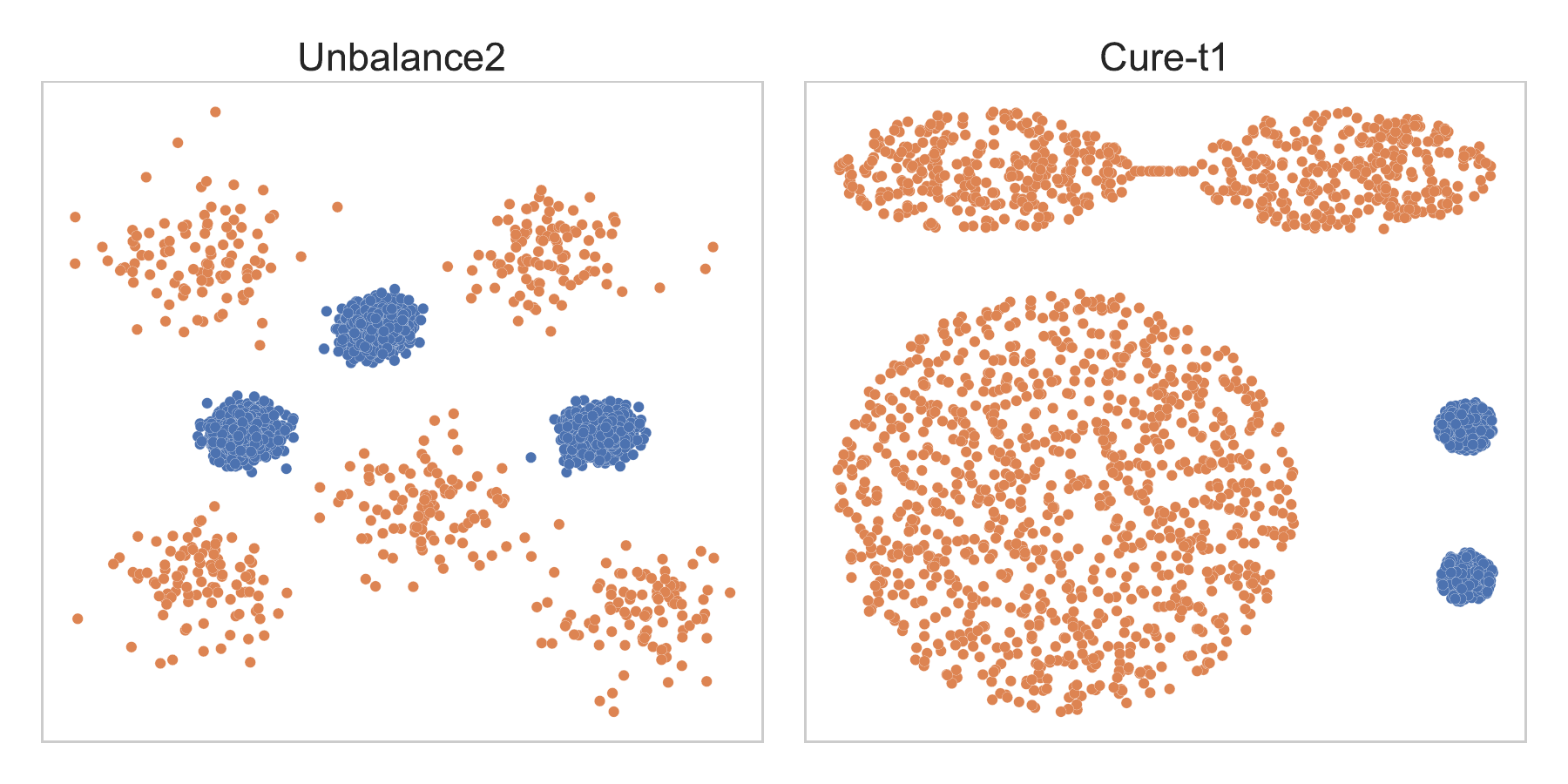}
        \end{minipage}%
    }
    \caption{The impact of hyperparameter combination on agent allocation. 
\textmd{(a) is the NMI results of \model{} under different parameter combinations, (b) is the information uncertainty distribution of intermediate nodes in the optimal two-level encoding tree, (c) is the agent allocation results on Unbalance2, Cure-t1, and Compound dataset with the parameter $Eps=0.3$ and $MinPts=1$. Each color corresponds to an agent.}}
    \label{fig:Hyperparameter agent allocation}
\end{figure*}

\subsubsection{Cluster based on Encoding Tree}\label{subsubsec:cluster based on encoding tree}
In our framework, there are three models that can be used to cluster datasets: \model{}, optimal two-level encode tree, and the submodel DRL-DBSCAN. 
Here, We experiment with the cluster accuracy of these three models and report the NMI results in Table~\ref{tab:cluster based on encoding tree}. 
The results of the optimal two-level encoding tree remain constant because the value of $k$ selected by Algorithm~\ref{alg:k selector} remains unchanged for the same dataset. 
The optimal two-level encode tree demonstrates strong performance in cluster applications as an unsupervised cluster model, even achieving the highest performance on the Compound and Pathbased datasets. 
As the overall model, \model{} shows superiority over other models on seven of the nine datasets. 
This demonstrates the overall effectiveness of the design, highlighting the effectiveness of its design in enhancing cluster accuracy and improving the robustness of the submodel.

\subsubsection{Hyperparameter on Agent allocation}\label{subsubsec:hyperparameter on agent allocation}
To pre-partition the dataset based on density, we calculate the information uncertainty of the intermediate nodes in the optimal encode tree and simply employ a DBSCAN to cluster the intermediate nodes based on information uncertainty. 
Here, we investigate the influence of different combinations of DBSCAN parameters on agent allocation and report the results in Figure~\ref{fig:Hyperparameter agent allocation}. 
Figure~\ref{fig:Hyperparameter agent allocation} (a) illustrates the impact of the parameter combinations on model performance. 
It shows the parameter $Eps$ has a negligible effect on the model, except for the D31 dataset. 
This is due to the fact that the clusters in the D31 dataset exhibit similar densities, resulting in equivalent levels of information uncertainty for intermediate nodes. 
Consequently, any adjustment to $Eps$ on the D31 dataset can lead to a significant alteration in the allocation of agents. 
On the other hand, the value of $Minpts$ is more significant since the optimal two-level encoding tree contains only a small number of intermediate nodes. 
Figure~\ref{fig:Hyperparameter agent allocation} (b) depicts that the information uncertainty proposed in Eq.~\ref{eq:information uncertainty} exhibits similar magnitudes across diverse datasets.
Figure~\ref{fig:Hyperparameter agent allocation} (c) shows the number of agents and the corresponding dataset partition for various datasets with the parameter combination of $Eps$=0.3 and $MinPts$=1.
It is evident that for the Unbalance2 and Cure-t1 datasets, the clusters with high density are assigned to one agent, while the clusters with low density are assigned to the other agent. 
This illustrates the effectiveness of our agent assignment method in accurately differentiating cluster densities.

\begin{figure}[t]
\centering
\includegraphics[width=0.8\textwidth]{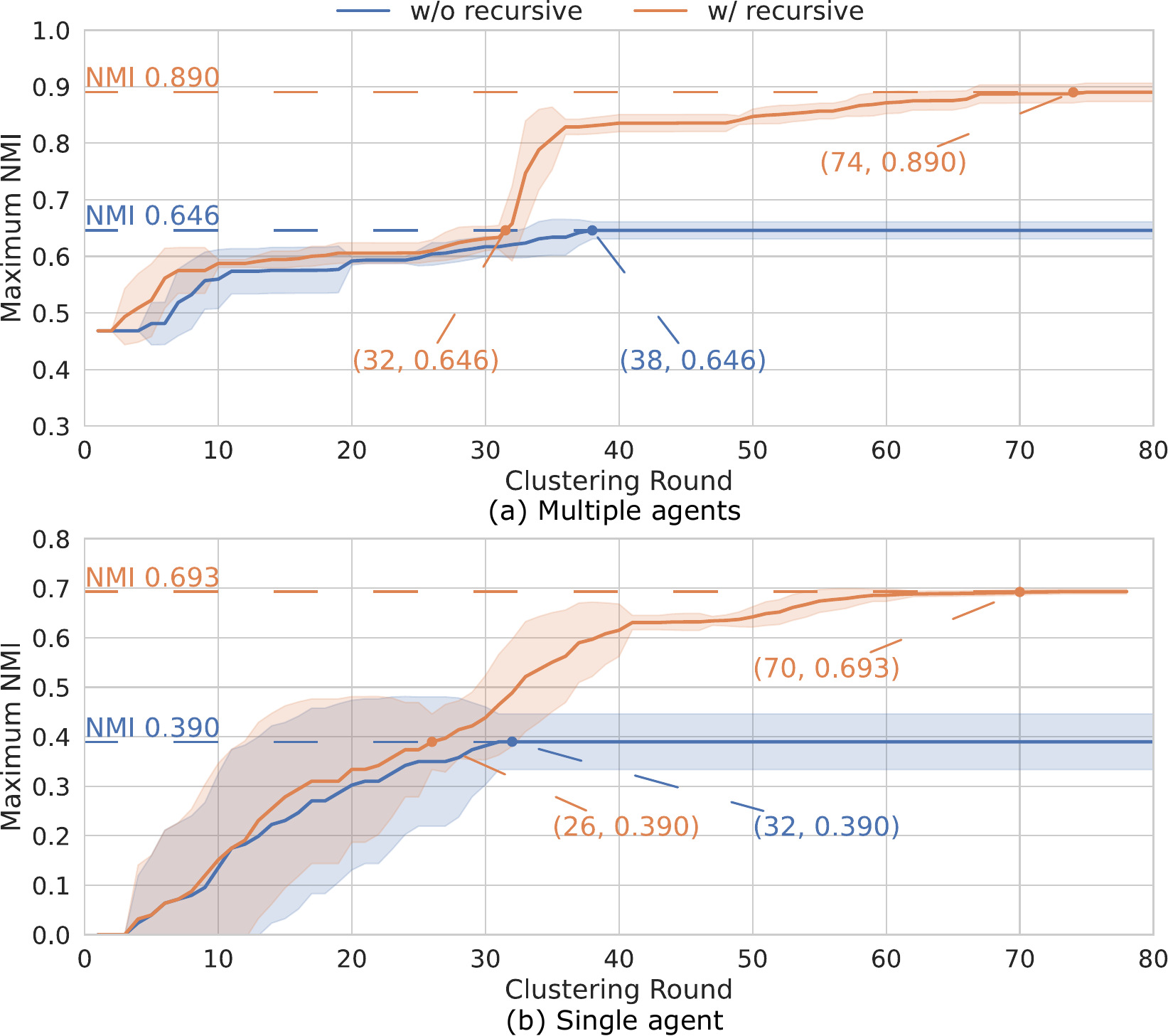}
\caption{The impact of recursive mechanism. \textmd{\textit{w/o recursive} in the legend denotes a model without using the recursive mechanism, while \textit{w/ recursive} indicates the utilization of the recursive mechanism.}}
\label{fig:recursive mechanism}
\end{figure}


\begin{figure*}[htb]
    \centering
    \includegraphics[width=0.8\linewidth]{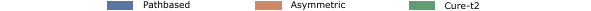}
    \subfigure[Eps Size ($\pi_{Eps}$)]{
        \begin{minipage}[t]{0.4\textwidth}
            \centering
            \includegraphics[width=1\linewidth]{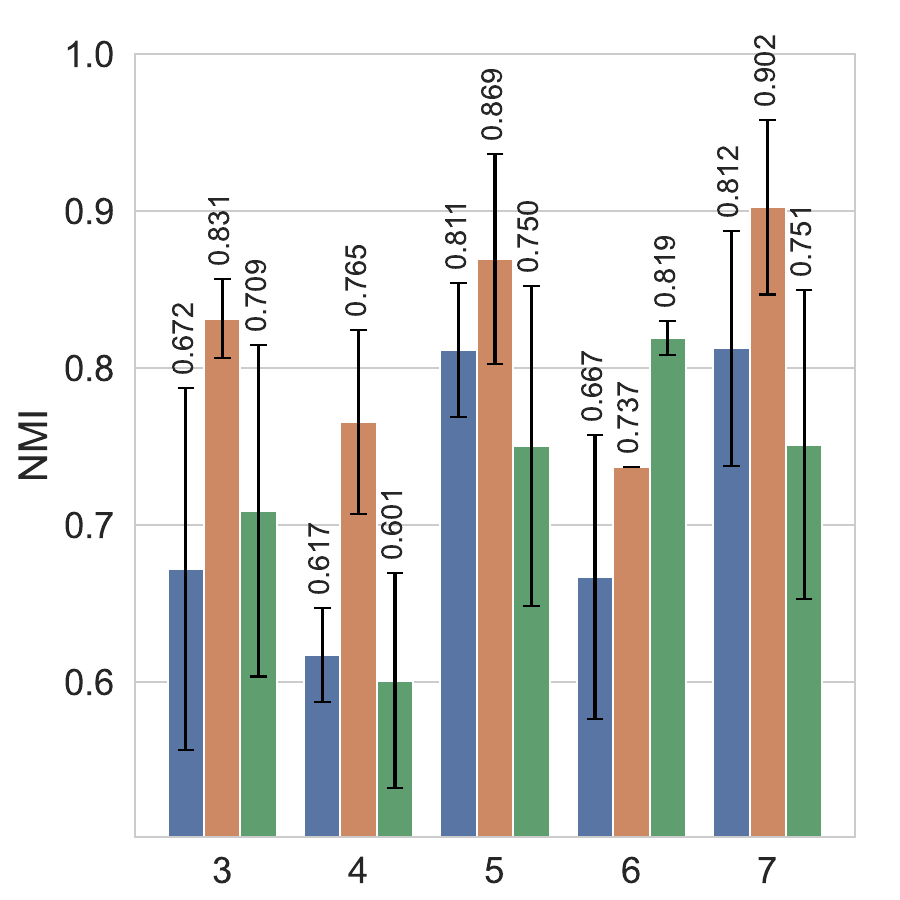}
        \end{minipage}%
    }
    \subfigure[MinPts Size ($\pi_{Minpts}$)]{
        \begin{minipage}[t]{0.4\textwidth}
            \centering
            \includegraphics[width=1\linewidth]{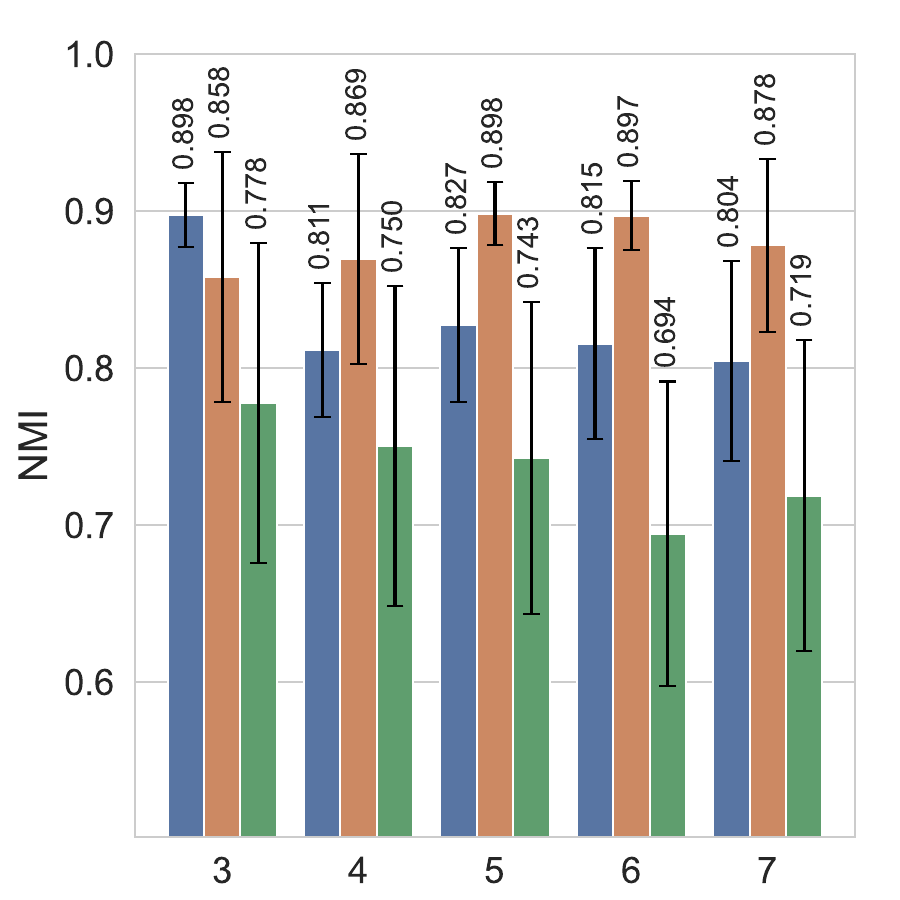}
        \end{minipage}%
    }
    \subfigure[Number of Layers ($l$)]{
        \begin{minipage}[t]{0.4\textwidth}
            \centering
            \includegraphics[width=1\linewidth]{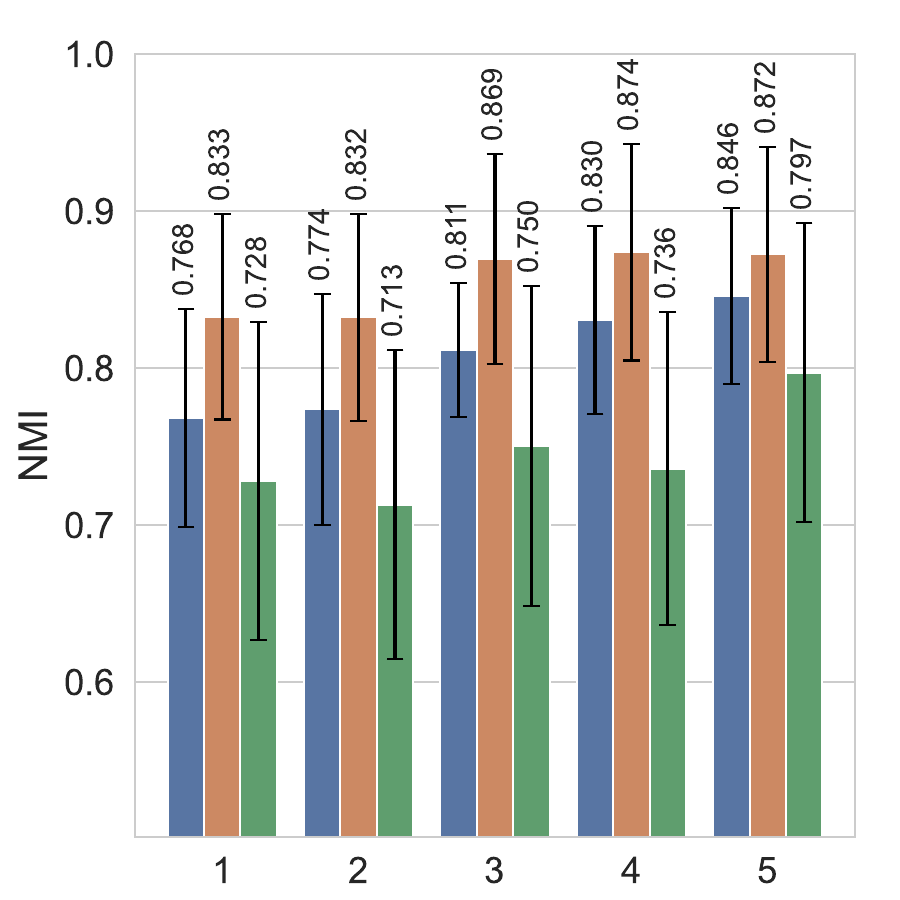}
        \end{minipage}%
    }
    \subfigure[Reward Factor ($\delta$)]{
        \begin{minipage}[t]{0.4\textwidth}
            \centering
            \includegraphics[width=1\linewidth]{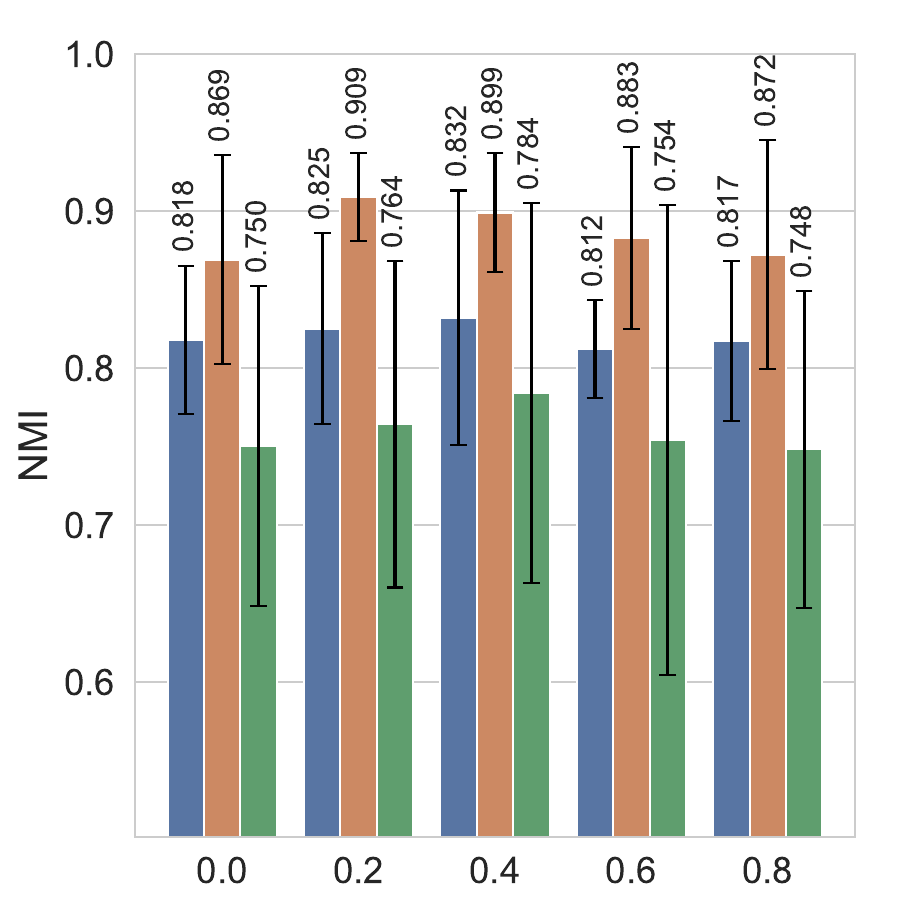}
        \end{minipage}%
    }
    \caption{Sensitivity analysis of the hyperparameters in the parameter search process.}
    \label{fig:hyperparameter on DRL}
\end{figure*}

\begin{figure*}
\centering
\includegraphics[width=0.97\textwidth]{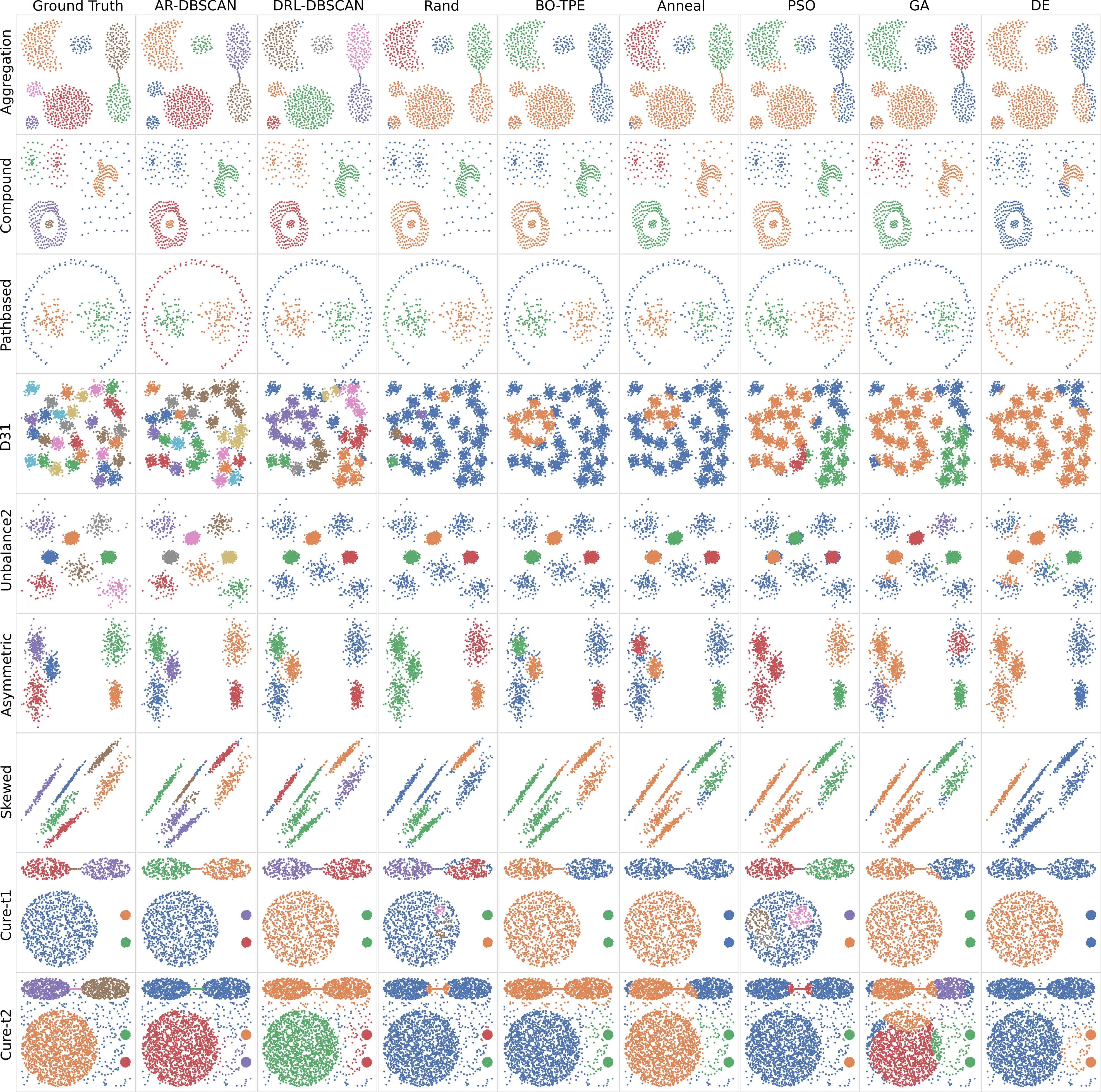}
\caption{The best results of multiple models across 30 clustering rounds. \textmd{The datasets organized from top to bottom: Aggregation, Compound, Pathbased, D31, Unbalance2, Asymmetric, Skewed, Cure-t1, and Cure-t2.}}
\label{fig:case study}
\end{figure*}

\subsubsection{Recursive Mechanism}\label{subsubsec:recursive mechanism}
We evaluate the impact of the recursive mechanism on the D31 dataset and report the maximum NMI achieved across clustering rounds in Figure~\ref{fig:recursive mechanism}. 
To better compare, we turn off the early stop mechanism so that \model{} can perform a longer search. 
In order to consider the influence of agent allocation on the overall search boundary for each agent, we conduct the experiment in scenarios involving both multiple agents and a single agent in Figure~\ref{fig:recursive mechanism} (a) and (b), respectively.
It can be observed that the utilization of the recursive mechanism can enhance the efficiency of parameter search by saving six rounds of search to reach the same peak as the variant without the recursive mechanism. 
Furthermore, we also observe that the recursive mechanism contributes to the improvement of the final convergence performance. 
This improvement can be attributed to the recursive mechanism narrowing down the search range and incrementally enhancing search precision layer by layer. 
Consequently, it mitigates the issue of becoming trapped in local optima, which can occur when employing a direct search with high precision.

\subsubsection{Hyperparameter on Parameter Search Process}\label{subsubsec:hyperparameter on parameter search process}
Figure~\ref{fig:hyperparameter on DRL} illustrates the results of \model{} on the Pathbased, Asymmetric, and Cure-t2 datasets with various hyperparameters. 
Figures~\ref{fig:hyperparameter on DRL} (a) and (b) show the effects of varying the parameter space size $\pi_{Eps}$ and $\pi_{MinPts}$. The results indicate that the space size of $Eps$ is a critical factor for model performance, and the optimal value varies for different datasets. 
The optimal value of $\pi_{Eps}$ for the Pathbased and Asymmetric datasets is 5, while 6 is more suitable for the Cure-t2 dataset. 
Contrastively, \model{} is less sensitive to $\pi_{MinPts}$ compared to $\pi_{Eps}$. 
Figure~\ref{fig:hyperparameter on DRL} (c) indicates that more layers generally lead to better model performance, as deeper recursion in parameter search space results in a more precise search. 
However, a higher number of layers may decrease model stability, as seen in the Asymmetric dataset, where setting the layer number $l$ to 3 achieves the best performance with greater stability.
Figure~\ref{fig:hyperparameter on DRL} (d) shows the impact of the reward factor $\delta$, which influences the weights of the endpoint immediate reward and future maximum immediate reward on the final reward (Eq.~\ref{eq:reward}). 
The results demonstrate that balancing the weight of the two immediate rewards generally benefits the model.

\subsection{Case Study}\label{subsec:case study}
To demonstrate the practicality and effectiveness of \model{}, we conduct a comprehensive case study and report the results in Figure~\ref{fig:case study}. 
Specifically, we evaluate \model{}, DRL-DBSCAN, Rand, BO-TPE, Anneal, PSO, GA, and DE across nine datasets and 30 clustering rounds. 
The figure illustrates the best clustering results obtained from each model.
It can be found that DRL-DBSCAN effectively distinguishes the three clusters located in the bottom left corner of the Aggregation dataset. 
\model{} showcases a similar capability, but it incorrectly assigns the two small clusters to the same cluster.
On the other hand, the other models mistakenly group them as a single cluster. 
DE performs the least satisfactorily, as it separates the seven-class dataset into two clusters. 
On the Compound dataset, \model{} is the only model that is able to accurately separate the dense cluster situated in the right half from the surrounding sparse cluster. 
This is attributed to the pre-partitioning approach we proposed based on information uncertainty, which allows for advanced separation of these two clusters. 
Other models struggle with clustering the dense cluster and part of the surrounding data points together, leading to inaccurate results. 
We also observe that all models fail to separate the two clusters located in the upper-left region, which are connected by a small neck. 
Most models perform well on the Pathbased dataset. 
However, on the D31 dataset, most models are only able to distinguish two clusters, while \model{} is capable of identifying a larger number of clusters.
Except for \model{}, all other models are affected by the three high-density clusters in the Unbalance2 dataset during the parameter search process, resulting in the misidentification of the other five sparse clusters as noise. 
In contrast, \model{} assigns an agent to the high-density clusters and the other agent to the low-density clusters. 
This allocation enables \model{} to achieve the nearly perfect distinction of all eight clusters.
In the case of the Asymmetric dataset, \model{} is the only model capable of separating all five clusters, while others incorrectly cluster some of the clusters together. 
The Skewed dataset is a challenging dataset for all models, as they all struggle with incorrectly dividing the end of certain oblong clusters into other clusters. 
Still, \model{} only misses one of the clusters completely, while other baselines miss at least 3 of the clusters.
On the Cure-t1 dataset, all models fail to separate the narrow bridges between the two elliptical clusters. 
\model{} successfully clusters all other clusters, while DRL-DBSCAN struggles to cluster two small-scale clusters into separate clusters. 
We attribute this to the $Eps$ value being excessively large, primarily influenced by the presence of two elliptical clusters and a large, sparse circular cluster situated in the lower-left region. 
The Cure-t2 dataset is generated by adding the same number of noise points as the data size of the Cure-t1 dataset. 
We observe that all models fail to correctly separate the two elliptical clusters in this dataset. 
Nonetheless, \model{} still recognizes the largest number of clusters correctly.

\section{Related Work}\label{sec:relate}

\noindent
\textbf{Structural Entropy.} 
Unlike early information entropy, such as Shannon entropy, which is defined by unstructured probability distributions, structural entropy~\cite{li2016structural} takes the hierarchical structural information of the input data into account.
It defines the encoding tree and structural entropy to evaluate the complexity of the hierarchical structure graph-structured data.
The process of constructing and optimizing the encoding tree is also a natural vertices clustering method for graphs.
Due to the theoretical completeness and interpretability of structural entropy theory, it has great potential for application in graph analyses such as graph hierarchical pooling~\cite{wu2022structural} and graph structure learning~\cite{zou2023se,duan2024structural}.
Moreover, the two-dimensional and three-dimensional structural entropy, which measure the complexity of hierarchical structures at two and three dimensions, respectively, have found applications in fields such as medicine~\cite{li2016three}, bioinformatics~\cite{li2018decoding}, social bot detection~\cite{pengunsupervised2024,zeng2024adversarial}, network security~\cite{li2016resistance}, and reinforcement learning ~\cite{zeng2023effective,zeng2023hierarchical}.

\noindent
\textbf{Automatic DBSCAN parameter determination.}
The clustering results of DBSCAN are heavily dependent on the settings of two sensitive hyperparameters, $Eps$ and $MinPts$, which are usually determined previously by practical experience. 
Numerous methods were proposed to address the challenge above.
OPTICS~\cite{ankerst1999optics} obtains the $Eps$ by establishing cluster sorting based on reachability, but the acquisition of $Eps$ needs to interact with the user.
V-DBSCAN~\cite{liu2007vdbscan} and KDDClus~\cite{mitra2011kddclus} plot a curve based on the sorted distances of each object to its $k$-th nearest neighbor and identify significant changes in the curve to generate a series of candidate values for the $Eps$ parameter. 
However, these methods fail to determine an appropriate value for the $MinPts$ parameter automatically.
Despite the advancements in clustering algorithms such as DSets-DBSCAN~\cite{hou2016dsets}, Outlier-DBSCAN~\cite{akbari2016automated}, and RNN-DBSCAN~\cite{bryant2017rnn}, these methods still rely on fixed $MinPts$ values or predetermined k-nearest neighbors~\cite{bryant2017rnn}. 
In addition to the approaches above, there have been efforts to identify the density of raw data points based on the size and shape of each data region determined by pre-defined grid partition parameters~\cite{darong2012grid, diao2018improved}. 
While these methods alleviate the challenge of parameter selection to some extent, they still rely on heuristic decisions from the user, limiting their adaptability to changing data.
Another viable method for parameter selection is based on the Hyperparameter Optimization (HO) algorithm.
BDE-DBSCAN~\cite{karami2014Choosing} aims to optimize an external purity index by employing a binary differential evolution algorithm to select the $MinPts$ parameters and a tournament selection algorithm to determine the $Eps$ parameters.
MOGA-DBSCAN proposes the outlier index as an internal index~\cite{falahiazar2021determining}. 
This index is utilized as the objective function, and a multi-objective genetic algorithm is employed to optimize and select the parameters.
AMD-DBSCAN~\cite{wang2022amd} proposes an improved parameter adaptation method within the algorithm to search for multiple parameter pairs for DBSCAN.
Only one easily defined hyperparameter is introduced to judge whether the clustering results are stable.
MDBSCAN~\cite{QIAN2024127329} divides data points into low-density parts and high-density parts, using a divide-and-conquer method to handle the respective parts to avoid their interference with each other.
However, expert knowledge is still required to manually determine the $k$-NN parameter and the relative density threshold parameter.
BAT-DBSCAN~\cite{BAT-DBSCAN} utilizes the BAT algorithm to automatically identify DBSCAN parameters while maximizing the purity function of the clustering.
However, it still encounters a performance drop when facing multi-density datasets.

\noindent
\textbf{Reinforcement Learning Clustering.}
Recently, Reinforcement learning (RL) has shown significant advantages in achieving adaptability through continuous interaction between agent and environment, and some works that integrate RL and clustering algorithms have been proposed.
AC-DRL~\cite{sharif2021dynamic} proposes an experience-driven approach based on an Actor-Critic based DRL framework to efficiently select the cluster head for managing the resources of the network, aiming to solve the challenges of the noisy nature of the Internet of Vehicles environment.
In the particle physics task, MCTS Clustering~\cite{brehmer2020hierarchical} utilizes Monte Carlo tree search to construct high-quality hierarchical clusters.
In the health and medical domain, \cite{grua2018exploring} proposes three distance metrics based on the state of the users and leverages two clustering algorithms and RL to cluster users who exhibit similar behaviors.
The methods mentioned above are field-specific RL clustering methods, but they cannot implement a general clustering framework.

\section{Conclusion}\label{sec:conclution}
In this paper, we proposed an adaptive and robust DBSCAN with the multi-agent reinforcement learning cluster framework, \model{}. 
The framework consists of two main components: a cluster density-based agent allocation method developed from the structure entropy theory and a deep reinforcement learning based parameter search method. 
The cluster density-based agent allocation method is designed to pre-partition the dataset based on the cluster density, which mitigates the problem of DBSCAN parameter search in varying density clusters. 
The deep reinforcement learning based parameter search method is applied to each agent for sensing the clustering environment and searching for the optimal parameter through weak supervision and an attention mechanism. 
A recursive search mechanism is devised to avoid the search performance decline caused by a large parameter space.
The experimental results on various tasks and datasets demonstrate the high accuracy, efficiency, and stability of our model. 
In future work, we plan to explore additional concepts for the interaction among multi-agents to avoid the impact of mis-partitioning on the agent allocation method.

\section*{Acknowledgments}
This work is supported by NSFC through grants 62322202, 62441612, 62432006, and 62476163.

\bibliographystyle{elsarticle-num-names}
\bibliography{references}

\appendix
\section{Selected $k$ in structured graph construction}\label{sec:selected k}
\begin{table*}[htp]
\aboverulesep=0ex
\belowrulesep=0ex
    \setlength{\tabcolsep}{0.9mm}
    \centering
    \caption{Selected $k$ for each dataset.}
    \begin{tabular}{ccccccccc}
    \toprule
        Aggregation & Compound & Pathbased & D31 & Unbalance2 & Asymmetric & Skewed & Cure-t1 & Cure-t2\\
        \hline
        22 & 8 & 5 & 9 & 1568 & 78 & 36 & 85 & 18\\
        \bottomrule
    \end{tabular}
    \label{tab:k selector}
\end{table*}
As stated in Sec. III-A1, \model{} first searches for $k$ to obtain the optimal $k$-NN graph.
We report the best $k$ selected for structured graph construction in Table~\ref{tab:k selector}.

\end{document}